\documentclass[10pt,twocolumn,letterpaper]{article}

\usepackage[pagenumbers]{cvpr} % To force page numbers, e.g. for an arXiv version
\usepackage{xcolor}         % colors
\usepackage{float}
\usepackage{placeins} 
\usepackage{cuted}
\usepackage{caption}
\usepackage{graphicx}
\usepackage{multirow}
\usepackage{amsmath}
\usepackage{tablefootnote}
\usepackage[table]{xcolor}
\usepackage{colortbl}
\usepackage{ragged2e}
\definecolor{cvprblue}{rgb}{0.21,0.49,0.74}
\usepackage[pagebackref,breaklinks,colorlinks,allcolors=cvprblue]{hyperref}
\usepackage{svg}
\def\paperID{44200} % *** Enter the Paper ID here
\def\confName{CVPR}
\def\confYear{2026}

\title{Context-Aware Semantic Segmentation via Stage-Wise Attention}

\author{
\parbox[t]{0.28\linewidth}{\centering
Antoine Carreaud\\
EPFL \& HEIG-VD\\
{\tt\small antoine.carreaud@epfl.ch}}
\and
\parbox[t]{0.28\linewidth}{\centering
Elias Naha\\
EPFL \& HEIG-VD\\
{\tt\small elias.naha@epfl.ch}}
\and
\parbox[t]{0.28\linewidth}{\centering
Arthur Chansel\\
EPFL \& HEIG-VD\\
{\tt\small arthur.chansel@epfl.ch}}
\and
\parbox[t]{0.28\linewidth}{\centering
Nina Lahellec\\
EPFL \& HEIG-VD\\
{\tt\small nina.lahellec@epfl.ch}}
\and
\parbox[t]{0.28\linewidth}{\centering
Jan Skaloud\\
EPFL\\
{\tt\small jan.skaloud@epfl.ch}}
\and
\parbox[t]{0.28\linewidth}{\centering
Adrien Gressin\\
HEIG-VD\\
{\tt\small adrien.gressin@heig-vd.ch}}\\
}

\begin{document}
\maketitle

\begin{abstract}
Semantic ultra-high-resolution (UHR) image segmentation is essential in remote sensing applications such as aerial mapping and environmental monitoring. Transformer-based models remain challenging in this setting because memory grows quadratically with the number of tokens, limiting either spatial resolution or contextual scope. We introduce \textbf{CASWiT} (Context-Aware Stage-Wise Transformer), a dual-branch Swin-based architecture that injects low-resolution contextual information into fine-grained high-resolution features through lightweight stage-wise cross-attention. To strengthen cross-scale learning, we also propose a SimMIM-style pretraining strategy based on masked reconstruction of the high-resolution image. Extensive experiments on the large-scale FLAIR-HUB aerial dataset demonstrate the effectiveness of CASWiT. Under our RGB-only UHR protocol, CASWiT reaches 66.37\% mIoU with a SegFormer decoder, improving over strong RGB baselines while also improving boundary quality. On the URUR benchmark, CASWiT reaches 49.2\% mIoU under the official evaluation protocol, and it also transfers effectively to medical UHR segmentation benchmarks. Code and pretrained models are available at \nolinkurl{https://huggingface.co/collections/heig-vd-geo/caswit}.
\end{abstract}

\section{Introduction}
\label{sec:intro}

Semantic segmentation of remote sensing imagery is central to many geospatial applications, including land-use mapping, environmental monitoring, and disaster response. As these applications increasingly rely on ultra-high-resolution (UHR) aerial imagery, methods must preserve fine local structures while also exploiting broader spatial context.

Transformer-based architectures~\cite{ViT,swin,pvtv2,mask2former} have advanced state-of-the-art results in vision tasks, but their application to UHR inputs remains challenging because of quadratic complexity and GPU memory constraints. Common workarounds such as downsampling or tiling either reduce useful context or sacrifice spatial resolution, impairing segmentation at both object and scene levels. Recent UHR approaches therefore combine high-resolution patch processing with explicit context modeling via multi-branch or cross-scale designs~\cite{Chen2021,boosting,GPWFormer}.

\paragraph{Our approach.}
\textbf{CASWiT}, a dual-branch hierarchical transformer, is introduced for RGB-only UHR segmentation. One branch processes high-resolution (HR) crops to preserve boundaries and small objects, while a second branch ingests wider low-resolution (LR) patches to encode global context. The two streams interact at multiple encoder stages through compact global cross-attention blocks (HR queries over LR keys/values), enabling early context injection while remaining compute-efficient. To further improve cross-scale learning, we adapt SimMIM-style~\cite{SimMIM} masked image modeling to this dual-stream setting and pretrain on large amounts of unlabeled orthophotos.

\paragraph{Benchmarks.}
The primary evaluation is conducted on FLAIR-HUB~\cite{flairhub} using an RGB-only UHR protocol that exploits its geospatial structure to reconstruct large contiguous tiles with preserved long-range context. Compared with URUR~\cite{WSDNET}, FLAIR-HUB provides a larger-scale and more carefully curated benchmark for cross-scale learning (see \S~\ref{sec:datasets}), while URUR is retained for continuity with prior work. We also report transfer results on medical image segmentation benchmarks to assess whether the proposed design extends beyond remote sensing.

\paragraph{Contributions.}
\begin{itemize}
  \item We introduce CASWiT (Context-Aware Stage-Wise Transformer), a dual-branch architecture that injects LR context into HR features through stage-wise cross-attention while preserving fine-grained HR detail.
  \item We design a dual-stream SimMIM pretraining strategy that strengthens cross-scale learning and transfers effectively to large-scale UHR segmentation tasks.
  \item We establish an RGB-only UHR evaluation protocol on FLAIR-HUB and show consistent improvements over prior RGB-only state of the art on FLAIR-HUB-RGB and URUR, with additional transfer results on medical segmentation benchmarks.
\end{itemize}

\section{Related Work}\label{sec:related_work}

\paragraph{Dual-stream UHR segmentation.}
Processing ultra-high-resolution (UHR) imagery for semantic segmentation requires preserving fine details while aggregating long-range context. A widely adopted strategy is dual-stream fusion, with an HR stream for local structures and an LR/context stream for scene-level semantics. GLNet~\cite{GLNET} popularized this formulation with CNN backbones and late fusion by concatenation. Subsequent works refine this template through alternative fusion schemes, backbones, or efficiency-oriented designs, including WSDNet~\cite{WSDNET}, FCtL~\cite{FcTL}, GPWFormer~\cite{GPWFormer}, SGNet~\cite{SGNet}, STUNet~\cite{STUNet}, and DESformer~\cite{DESFormer}. Other UHR methods target iterative patching with global guidance, proposal-based computation, shallow all-pixel processing, or boundary refinement~\cite{PPN, LDNET, ISDNET, WSDNET, GPWFormer, SGHRQ, WCTNet, EFFNet, RingFormerSeg, CascadePSP, Magnet, FcTL}. Most of these approaches still rely on mid/late fusion, where HR and LR features interact only after substantial single-stream processing, although earlier interaction can be beneficial when the two inputs share similar representational spaces~\cite{Baltrusaitis2019Survey}.

\paragraph{Single-stream HR backbones.}
An alternative to dual-stream fusion is to rely on hierarchical vision transformers or multi-scale CNNs that capture locality and globality within a single stream. Representative examples include Swin Transformer~\cite{swin, swinv2} and PVT~\cite{pvt,pvtv2}, alongside UHR-oriented single-stream variants~\cite{CMTFNet,multi_scale_progressive,MSTRANS,MESTRANS,GLAM,MFNet}. Recent efficient transformer variants also aim to mitigate the quadratic cost of self-attention through architectural approximations or more scalable attention patterns~\cite{Linformer,Performer,FocalNet}. While these models improve scalability compared to vanilla ViT, balancing spatial resolution and memory for truly UHR inputs remains challenging.

\paragraph{Fusion mechanisms and module placement.}
Despite the intuitive complementarity of HR and LR signals, many dual-stream models still rely on concatenation or summation at mid or late stages. Attention-based alignment between heterogeneous resolutions is less explored across multiple depths, even though cross-attention provides a systematic way to condition HR features on LR context early in the hierarchy. DESformer~\cite{DESFormer} introduces a multi-depth feature interaction module inside the encoder, while CTCFNet~\cite{CTCFNet} combines CNN-Transformer features through a mid-to-late aggregation module and a bi-directional decoder. More recently, Boosting Dual-Stream~\cite{boosting} revisits the dual-stream paradigm with uncertainty-guided HR/LR interaction. Compared with these works, CASWiT uses two hierarchical transformer branches and lightweight stage-wise cross-attention from the earliest encoder stages, and further couples this design with SimMIM-style pretraining~\cite{SimMIM}.%This positioning is important in the UHR setting, where contextual cues must be injected early enough to influence fine-grained HR features without sacrificing scalability.

\paragraph{Other UHR dense prediction tasks.}
Similar mechanisms have also been explored in other dense UHR tasks such as salient object detection or monocular depth estimation, where dual-stream designs, patch selection, and uncertainty estimation remain important~\cite{ISDNET,PPN,ESNet_hongyu,remote_sensing_gated,deep_deterministic,sub_ensemble_uncertainty}. More broadly, the need to combine fine structures with wider context also arises beyond remote sensing, including medical image segmentation, which motivates our transfer experiments.

\begin{figure*}[!htb]
  \centering
  %\includegraphics[width=1\textwidth]{figures/architecture.png}
  %gauche bas droite haut
  \includegraphics[width=0.97\textwidth,trim=0cm 3cm 0cm 3.5cm,clip]{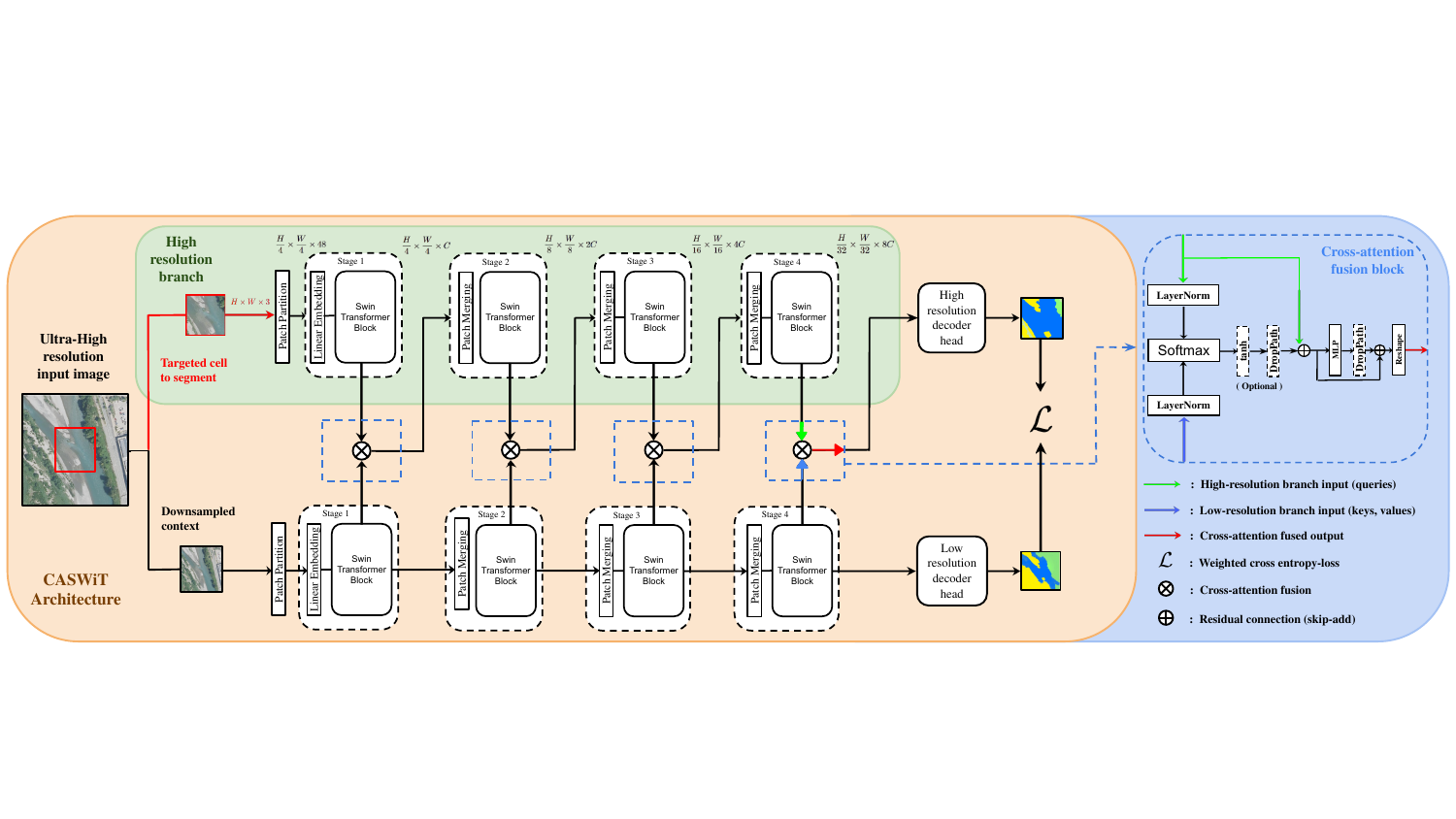}
  \caption{\textbf{CASWiT architecture.} A dual-branch encoder for ultra-high-resolution imagery: the HR branch processes the target tile, while the LR branch encodes a downsampled larger context. At each Swin stage (1$\rightarrow$4), HR features provide queries and LR features provide keys/values to a cross-attention module with residual fusion and an optional gate (right). HR and LR decoder heads are jointly optimized during training with a weighted cross-entropy loss.}
  \label{fig:arch}
\end{figure*}

\section{Method}\label{sec:method}

CASWiT (Context-Aware Stage-Wise Transformer) is a dual-branch architecture that fuses high-resolution (HR) features with low-resolution (LR) contextual features through compact cross-attention blocks inserted after each encoder stage (Fig.~\ref{fig:arch}). Each block applies HR-LR cross-attention followed by a residual MLP; we also consider an optional learned gate, but use the ungated variant by default. The network is trained with supervision on the HR output and an auxiliary LR loss weighted by $\alpha$. At inference, the LR stream remains to provide context, but its decoder/head is removed.

\subsection{Overview}
CASWiT combines a HR Swin encoder~\cite{swin} that preserves high resolution features with a LR Swin encoder that captures global contextual features from a larger field of view. Both encoders share identical hierarchical configurations (stages $\{1..4\}$, channel schedule $C_s$). Cross-attention modules are inserted after each stage to inject LR context into the HR stream. A UPerNet~\cite{upernet} or SegFormer~\cite{segformer} head processes the HR features to produce the final logits.

\subsection{Dual-resolution encoder}\label{subsec:encoder}

\paragraph{Inputs.}
Given an HR crop $I^{\mathrm{HR}}\!\in\!\mathbb{R}^{H\times W\times3}$ and a co-registered LR image $I^{\mathrm{LR}}$ (downsampled from a larger FoV), the two Swin encoders produce stage-wise feature maps:
\[
X^{\mathrm{HR}}_s\!\in\!\mathbb{R}^{H_s\times W_s\times C_s},\qquad
X^{\mathrm{LR}}_s\!\in\!\mathbb{R}^{\hat H_s\times\hat W_s\times C_s}.
\]
The HR and LR features may differ in spatial size; they are flattened into token sequences before fusion.

\paragraph{Cross-attention fusion block.}
At each stage $s$, we perform multi-head cross-attention (MHA) from HR queries to LR keys/values:
\[
Q = \mathrm{LN}(X^{\mathrm{HR}}_s)W_Q
\]
\[
K = \mathrm{LN}(X^{\mathrm{LR}}_s)W_K, \quad V = \mathrm{LN}(X^{\mathrm{LR}}_s)W_V,
\]
\[
A_s = \mathrm{MHA}\!\big(Q,\ K,\ V\big).
\]
The final HR features at stage $s$, $\tilde H_s$, are obtained via a residual connection and an optional learned gate $\gamma_s$:
\[
H'_s = X^{\mathrm{HR}}_s + \gamma_s \odot A_s,\qquad
\tilde H_s = H'_s + \mathrm{MLP}(H'_s),
\]
where $\gamma_s = \tanh(g_s)$ is a learned scalar stage-wise gate broadcast over HR tokens.  
The gate controls how much contextual information from the LR stream is injected into HR features. In practice, the ungated variant slightly outperforms the gated one, so we use the ungated version by default and report both settings in \S~\ref{sec:experiments}.

\subsection{Decoder and prediction heads}\label{subsec:decoder}
We consider two decoder variants for the HR branch: UPerNet~\cite{upernet} and SegFormer~\cite{segformer}. In the default setting, we adopt UPerNet, where stage features $\{\tilde X^{\mathrm{HR}}_s\}_{s=1}^4$ are fused by a Feature Pyramid Network (FPN) with a Pyramid Pooling Module (PPM) and then classified by the head. We also evaluate a SegFormer decoder on top of the same HR features. A LR decoder mirrors the same structure and is used only during training for auxiliary supervision; it can be removed at inference.

\subsection{Supervised objectives}\label{subsec:supervised}
Let $\hat Y^{\mathrm{HR}}\in\mathbb{R}^{H\times W\times K}$ and $\hat Y^{\mathrm{LR}}\in\mathbb{R}^{\hat H\times \hat W\times K}$ be the logits from the HR and LR heads, and let $Y\in\{1,\ldots,K\}^{H\times W}$ be the ground-truth labels. 
We compute standard pixel-wise cross-entropy on HR:
\[
\mathcal{L}_{\text{HR}} = -\frac{1}{HW}\sum_{p}\sum_{k}\mathbf{1}[Y_p{=}k]\,
\log \mathrm{Softmax}(\hat Y^{\mathrm{HR}}_p)_k.
\]
For LR supervision, we use the downsampled label map $Y^\downarrow\in\{1,\ldots,K\}^{\hat H\times \hat W}$ (nearest-neighbor):
\[
\mathcal{L}_{\text{LR}} = -\frac{1}{\hat H\hat W}\sum_{p}\sum_{k}\mathbf{1}[Y^\downarrow_p{=}k]\,
\log \mathrm{Softmax}(\hat Y^{\mathrm{LR}}_p)_k.
\]
The total loss is the weighted sum
\[
\mathcal{L} = \mathcal{L}_{\text{HR}} + \alpha\,\mathcal{L}_{\text{LR}},
\]
where $\alpha$ controls the contribution of the LR auxiliary head (set to $0.5$ in our experiments).

\subsection{Self-supervised pretraining (SimMIM-style)}\label{subsec:ssl}
We adapt a simple framework for masked image modeling (SimMIM)~\cite{SimMIM} to the dual-stream encoder and keep the HR-LR fusion active throughout pretraining.

\paragraph{Masking strategy.}
On the HR stream, we apply random masking with ratio $r_{\mathrm{HR}}$ (default $0.75$). On the LR stream, we apply centered masking with ratio $r_{\mathrm{LR}}$ (default $0.5$), masking the LR region that is spatially aligned with the HR crop. This preserves the surrounding global layout while preventing trivial cross-scale copying from the LR center to the HR target region. In both cases, the masked tokens are replaced with a learnable mask token at the stage 1 embedding dimension (with no zeroing), as in many frameworks~\cite{MAE, SimMIM}. Cross-attention therefore operates on LR features whose masked positions carry the learned mask token embedding.

\begin{figure*}
    \centering
    \includegraphics[width=1\linewidth]{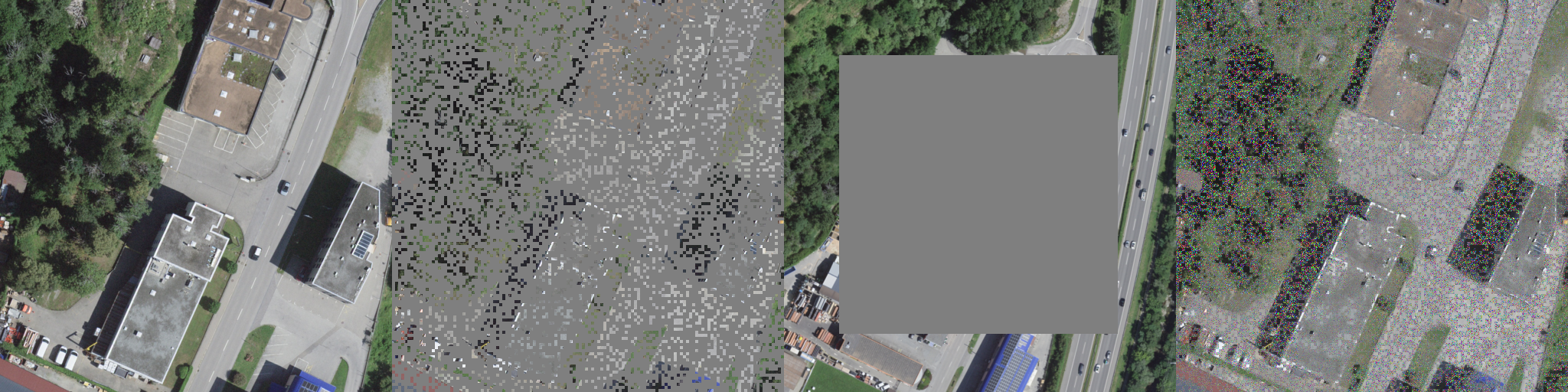}
  \caption{Self-supervised inference results on the CASWiT architecture. Each image (left to right) shows: original high-resolution image, high-resolution image with random masking, low-resolution image with central masking, and the reconstruction of the high-resolution image after SimMIM-style pretraining.}
  \label{fig:mae_figure}
\end{figure*}

\paragraph{Reconstruction head and objective.}
Only the HR branch is reconstructed. From the last HR stage ($s{=}4$), we use a $1{\times}1$ convolution producing $3s^2$ channels followed by a PixelShuffle with stride $s$ (equal to the total downsampling factor of the HR encoder) to map tokens back to RGB at input resolution. Let $\hat I^{\mathrm{HR}}$ be the reconstruction. We minimize a masked $\ell_1$ loss over the masked HR pixels only:
\[
\mathcal{L}_{\text{SSL}} \;=\; \frac{1}{3\,|M^{\text{pix}}_{\mathrm{HR}}|}\sum_{p\in M^{\text{pix}}_{\mathrm{HR}}} \big\| \hat I^{\mathrm{HR}}_p - I^{\mathrm{HR}}_p \big\|_1
\]
where $M^{\text{pix}}_{\mathrm{HR}}$ is obtained by upsampling the HR \emph{patch} mask to pixel resolution using the stage-1 patch size. During SSL, masked tokens are replaced by a learnable mask embedding; fusion remains active so the encoder can leverage LR semantics to infer missing HR content.
Fig.~\ref{fig:mae_figure} illustrates this dual-stream masking strategy, with random HR masking and centered LR masking during self-supervised pretraining. After SSL, we discard the reconstruction head and fine-tune the dual-stream encoders with cross-attention under the supervised objective in \S~\ref{subsec:supervised}.

\section{Experiments}\label{sec:experiments}
\subsection{Datasets}\label{sec:datasets}

\begin{figure}[htb]
  \centering
  \includegraphics[width=0.8\linewidth]{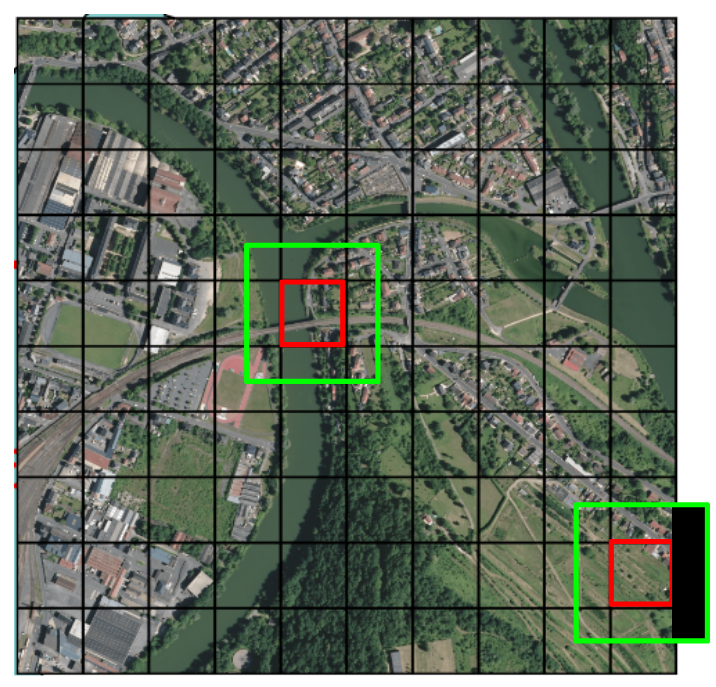}
  \caption{HR/LR construction on FLAIR-HUB. 
  Red: \,original HR patch (512$\times$512). 
  Green: \,georeferenced 3$\times$3 neighborhood assembled into a 1024$\times$1024 context, then downsampled $\times$2 to form the LR input (512$\times$512).}
  \label{fig:pre-process}
\end{figure}

\paragraph{FLAIR-HUB.}
We use the FLAIR-HUB dataset~\cite{flairhub}, a large-scale multimodal extension of FLAIR~\cite{FLAIR1}, comprising 241{,}100 RGB patches of size 512$\times$512 at 0.20\,m GSD, annotated into 15 classes. 
To enable RGB-only UHR evaluation while remaining comparable to the official per-patch setting, we construct for each HR patch a geospatially aligned 3$\times$3 context tile using its eight neighbors (GeoTIFF coordinates), yielding a 1024$\times$1024 composite that we downsample by 2 to obtain a 512$\times$512 LR input co-registered with the HR patch. 
When neighbors are missing at borders, we fill the gaps with black padding to keep the same dimensions for all patches.
This protocol preserves long-range spatial context while keeping the input size compatible with standard backbones (see Fig.~\ref{fig:pre-process}).

\paragraph{URUR.}
The URUR dataset~\cite{WSDNET} contains 3{,}008 UHR RGB images of size 5120$\times$5120 from 63 cities with 8 land-cover classes. 
We follow the official split: 2{,}157 train, 280 val, 571 test. 
URUR has been influential for UHR evaluation, however, we observed occasional image-mask inconsistencies (e.g., local misalignment) that can affect evaluation metrics. 
We therefore report URUR results, but we advise interpreting URUR metrics with care: scores can be underestimated or display higher variance due to occasional image-mask non-conformities (illustrative examples are provided in the supplementary, Fig.~\ref{fig:urur_mismatch_supp}), and, as discussed in~\cite{boosting}, the handling of the other class may depress IoU (near zero) because it appears sparsely.

\paragraph{SWISSIMAGE (unlabeled, for SSL).}
For self-supervised pretraining, we use large-scale unlabeled orthophotos from the SWISSIMAGE archive at 0.20\,m GSD (total ${\sim}$1067\,Gpx; excluded from supervised splits). 
This corpus provides over 40$\times$ more pixels than labeled training data from official test split of FLAIR-HUB, enabling robust masked reconstruction pretraining.

\paragraph{Other UHR benchmarks.}
For completeness, we note that smaller remote sensing benchmarks such as INRIA Aerial~\cite{inria} and DeepGlobe~\cite{Demir2018} are also commonly reported, but we focus on FLAIR-HUB and URUR because they provide a stronger setting for large-scale multi-class cross-scale evaluation. To assess transfer beyond aerial imagery, we additionally evaluate CASWiT on two medical UHR benchmarks, ISIC~\cite{isic} and CRAG~\cite{CRAG_dataset}.

%\paragraph{Other UHR benchmarks.}
%For completeness, we note that the community frequently reports on remote sensing benchmarks such as INRIA Aerial~\cite{inria} and DeepGlobe~\cite{Demir2018}. We do not include them in our main evaluation because (i) INRIA Aerial provides building-background annotations only (single-class target), which is less effective for multi-class UHR segmentation, and (ii) both datasets are significantly smaller than FLAIR-HUB and URUR, offering limited coverage for large-scale cross-scale analysis. We therefore focus our primary remote sensing study on FLAIR-HUB (primary, RGB-only UHR protocol) and URUR (legacy benchmark) to balance scale, class diversity, and continuity with prior work. To assess whether CASWiT transfers beyond aerial imagery, we evaluate it on two medical UHR segmentation benchmarks: ISIC~\cite{isic} and CRAG~\cite{CRAG_dataset}. ISIC is a dermoscopic lesion segmentation benchmark, while CRAG focuses on colorectal gland segmentation in histopathology. We use them as complementary high-resolution testbeds to evaluate the transferability of our dual-stream design beyond aerial orthophotos.

\subsection{Evaluation Protocols}
We report standard semantic segmentation metrics, including mean Intersection-over-Union (mIoU) and mean F1 score (mF1), computed over all non-void classes. 
For FLAIR-HUB, we follow the official split named "split\_flairhub" and report per-class results in the supplementary. There is no geographic overlap between train/val/test splits. To avoid patch-boundary bias when using geospatially reconstructed neighborhoods, while remaining directly comparable to the original patch-based FLAIR-HUB protocol, evaluation is performed only on the center crop corresponding to the original HR patch.
For URUR, we follow the original train/val/test protocol with the configuration of 8 classes (with class "other" for comparison with previous work) and also report per-class results in the supplementary.
For ISIC and CRAG, we use the official splits and report mIoU on the test sets.
Inferences are performed without overlapping sliding windows on FLAIR-HUB, URUR, ISIC and CRAG.

\paragraph{Mean Boundary IoU (mBIoU).}
Beyond region overlap, we evaluate boundary quality using the mean Boundary IoU metric (mBIoU)~\cite{boundary}. For each class $c$, we extract thin boundary bands from the ground truth ($B_{\hat{Y}}^c$) and the prediction ($B_Y^c$) by dilating their contours.

The boundary IoU for class $c$ and mBIoU are:
\[
\mathrm{bIoU}(c)=\frac{|B_Y^c \cap B_{\hat{Y}}^c|}{|B_Y^c \cup B_{\hat{Y}}^c|},
\quad
\mathrm{mBIoU}=\frac{1}{C}\sum_{c=1}^{C}\mathrm{bIoU}(c)
\]
Compared to standard mIoU, mBIoU is insensitive to large homogeneous regions and focuses on how well object edges are localized.

\subsection{Implementation Details}
All experiments are implemented in PyTorch and trained on 4$\times$~NVIDIA L40S GPUs (48\,GB each) using Distributed Data Parallel (DDP). 
We use the AdamW optimizer with an initial learning rate of \(6\times10^{-5}\), decayed to \(1\times10^{-6}\) through a cosine annealing scheduler, and a weight decay of \(0.01\). 
Batch size is set to 20 (5 per GPU) for URUR and 16 (4 per GPU) for FLAIR-HUB. 
Training runs for 20 epochs with a crop size of 512$\times$512 for both HR and LR inputs (LR initially 1024$\times$1024 and subsampled to 512$\times$512). 
No data augmentation is applied, unless specified, to ensure a controlled comparison across methods and datasets. 

For all experiments, both HR and LR branches use identical backbones, CASWiT-Base means the use of 2 Swin-B backbones. 
Unless otherwise specified, the gating mechanism is disabled (see ablation in \S~\ref{sec:ablation}), and the auxiliary LR supervision weight is set to \(\alpha = 0.5\). 

For capacity-controlled comparisons, we additionally evaluate a late-fusion dual-branch baseline with the same HR/LR inputs (using a simple sum). On the same dual-stream backbone, we report results for both decoder variants, UPerNet and SegFormer, and provide GFLOPs and inference speed (FPS) on FLAIR-HUB inputs.

\paragraph{Self-supervised pretraining.}
We perform SimMIM-style pretraining on the unlabeled SWISSIMAGE corpus for 100 epochs before fine-tuning. 
Masking ratios are set to 75\% for HR (random) and 50\% for LR (centered), the latter designed to maintain global layout while preventing trivial pixel copying across scales. 
Both streams and cross-attention blocks are optimized jointly during pretraining. 
The pretrained weights are then used for direct fine-tuning on FLAIR-HUB, URUR, ISIC and CRAG without intermediate adaptation.

\subsection{Quantitative Results}\label{sec:quantitative_results}

\paragraph{FLAIR-HUB (RGB-only UHR protocol).}
Table~\ref{tab:flairhub_results} reports RGB-only segmentation performance on FLAIR-HUB under the proposed UHR protocol. Compared with the four official Swin+UPerNet RGB baselines released by the dataset authors~\cite{flairhub}, CASWiT consistently improves performance: CASWiT-Base + UPerNet reaches 65.11\% mIoU, which further increases to 65.35\% with self-supervised pretraining and to 65.83\% with additional spatial and radiometric augmentations. Replacing UPerNet with SegFormer on the same CASWiT backbone yields the best result, \textbf{66.37\%} mIoU and \textbf{78.58\%} mF1, while also reducing compute from 489 to 298 GFLOPs and increasing inference speed from 15.4 to 17.9 FPS.

For reference, the best multimodal configuration reported in~\cite{maestro} reaches 65.9\% mIoU, meaning that our best RGB-only variant surpasses this score while relying solely on RGB imagery. CASWiT also improves boundary quality, with mBIoU increasing from 32.57 for the retrained Swin-B + UPerNet baseline to 35.87 for CASWiT-Base + UPerNet and 36.90 for CASWiT-Base-SSL-aug + UPerNet.

To isolate the effect of stage-wise cross-attention from model capacity, we additionally compare against a dual-branch late-fusion baseline with the same HR/LR inputs and decoder, as well as against a larger single-stream Swin-L baseline. The late-fusion variant reaches 64.25 mIoU, remaining below CASWiT despite using the same dual-stream setting. As an additional reproducible reference, retrained ISDNet reaches 52.77 mIoU on FLAIR-HUB-UHR, remaining well below all CASWiT variants.

\begin{table}[t]
\centering
%tiny
\scriptsize
\begin{tabular}{lccccc}
\toprule
\textbf{Model} & \textbf{mIoU} & \textbf{mF1} & \textbf{mBIoU} & \textbf{GFLOPs} & \textbf{FPS} \\
\midrule
\multicolumn{6}{l}{\emph{Official RGB baselines}} \\
Swin-T + UPer~\cite{flairhub} & 62.01 & 75.27 & - & 237 & 69.2 \\
Swin-S + UPer~\cite{flairhub} & 61.87 & 75.11 & - & 261 & 41.5 \\
Swin-B + UPer~\cite{flairhub} & 64.05 & 76.88 & - & 306 & 36.2 \\
\quad $\hookrightarrow$ retrained & 64.02 & 76.64 & 32.57 & 306 & 36.2 \\
Swin-L + UPer~\cite{flairhub} & 63.36 & 76.35 & - & 420 & 27.8 \\
\midrule
\multicolumn{6}{l}{\emph{Capacity-controlled}} \\
ISDNet~\cite{ISDNET} (retrained) & 52.77 & - & - & - & - \\
Dual Swin-B (late fusion) & 64.25 & - & - & 398 & 19.4 \\
\midrule
\multicolumn{6}{l}{\textbf{CASWiT (Ours)}} \\
Base + UPer & 65.11 & 77.71 & 35.87 & 489 & 15.4 \\
Base-SSL + UPer & 65.35 & 77.87 & 35.99 & 489 & 15.4 \\
Base-SSL-aug + UPer & 65.83 & 78.22 & \textbf{36.90} & 489 & 15.4 \\
Base-SSL-aug + SegF & \textbf{66.37} & \textbf{78.58} & 36.51 & \textbf{298} & \textbf{17.9} \\
\bottomrule
\end{tabular}
\caption{FLAIR-HUB test results under the RGB-only UHR protocol. CASWiT improves over the official RGB baselines and a capacity-controlled late-fusion variant. The SegFormer head yields the best mIoU and mF1 while reducing GFLOPs. UPer = UPerNet, SegF = SegFormer.}
\label{tab:flairhub_results}
\end{table}

\paragraph{URUR (legacy benchmark).}
Table~\ref{tab:urur_results} reports results on URUR. 
CASWiT improves over prior UHR-specific architectures such as WSDNet~\cite{WSDNET} and the recent Boosting Dual-Stream model~\cite{boosting}. It reaches 48.7\% mIoU with CASWiT-Base + UPer, 49.1\% with CASWiT-Base-SSL-aug + UPer, and 49.2\% with CASWiT-Base-SSL-aug + SegF.
These results support the benefit of stage-wise cross-attention for combining fine detail and large-scale context in UHR segmentation. 
As discussed in \S~\ref{sec:datasets}, occasional annotation inconsistencies and the handling of the \textit{other} class can lead to underestimated mIoU on URUR.

\begin{table}[t]
\centering
\scriptsize
\setlength{\tabcolsep}{6pt}
\begin{tabular}{lcc}
\toprule
\textbf{Model} & \textbf{mIoU} & \textbf{Mem (MB)} \\
\midrule
\multicolumn{3}{l}{\emph{Generic baselines}} \\
PSPNet~\cite{CascadePSP} & 32.0 & 5482 \\
ResNet18~\cite{ResNet} + DeepLabv3+~\cite{Deeplabv3+} & 33.1 & 5508 \\
STDC~\cite{STDC} & 42.0 & 7617 \\
\midrule
\multicolumn{3}{l}{\emph{UHR methods}} \\
GLNet~\cite{GLNET} & 41.2 & 3063 \\
FCtL~\cite{FcTL} & 43.1 & 4508 \\
ISDNet~\cite{ISDNET} & 45.8 & 4920 \\
WSDNet~\cite{WSDNET} & 46.9 & 4510 \\
Boosting Dual-Stream~\cite{boosting} & 48.2 & 3682 \\
CASWiT-Base + UPer & 48.7 & 2996 \\
CASWiT-Base-SSL-aug + UPer & 49.1 & 2996 \\
CASWiT-Base-SSL-aug + SegF & \textbf{49.2} & 2878 \\
\bottomrule
\end{tabular}
\caption{URUR test results. CASWiT improves over prior UHR-specific methods while remaining memory-efficient. UPer = UPerNet, SegF = SegFormer.}
\label{tab:urur_results}
\end{table}

\paragraph{Generalization beyond remote sensing: ISIC and CRAG.}
To assess whether CASWiT transfers beyond aerial imagery, we evaluate it on two medical UHR segmentation benchmarks: ISIC~\cite{isic} and CRAG~\cite{CRAG_dataset}. Table~\ref{tab:medical_results} shows that CASWiT generalizes well without architectural modification, outperforming recent dual-branch UHR baselines on ISIC and matching or improving upon them on CRAG. Using a SegFormer decoder further improves performance, reaching 86.5 mIoU on ISIC and 90.7 on CRAG. These results indicate that the proposed stage-wise context injection is not limited to aerial imagery and transfers effectively to other high-resolution segmentation domains.

\begin{table}[t]
\centering
\scriptsize
\setlength{\tabcolsep}{8pt}
\begin{tabular}{l c c}
\toprule
\textbf{Model} & \textbf{ISIC (mIoU)} & \textbf{CRAG (mIoU)} \\
\midrule
GPWFormer~\cite{GPWFormer} & 80.7 & 89.9 \\
Boosting Dual-Stream~\cite{boosting} & 83.4 & 90.3 \\
\midrule
CASWiT-Base-SSL-aug + UPer & 85.4 & 90.3 \\
CASWiT-Base-SSL-aug + SegF & \textbf{86.5} & \textbf{90.7} \\
\bottomrule
\end{tabular}
\caption{Results on the ISIC and CRAG test sets. CASWiT transfers effectively beyond remote sensing and benefits further from the SegFormer head. UPer = UPerNet, SegF = SegFormer.}
\label{tab:medical_results}
\end{table}

\subsection{Ablation Studies}\label{sec:ablation}
We perform a series of ablation experiments on the FLAIR-HUB validation split to analyze the impact of the key design choices in CASWiT. 
Unless otherwise stated, all variants use a Swin-B backbone and are trained under identical conditions (20 epochs, crop size 512$\times$512, no data augmentation). 
We systematically vary the cross-attention pattern, the auxiliary LR supervision weight $\alpha$, the gating mechanism, and the SSL initialization. 
Results are summarized in Table~\ref{tab:ablation}.

\paragraph{Cross-attention.}
To isolate the effect of cross-scale fusion, we remove the LR/context branch and disable all cross-attention modules. 
This effectively reduces CASWiT to a standard Swin-B encoder with a UPerNet decoder, i.e., the RGB baseline used in the FLAIR-HUB paper. 
On the validation set, this single-stream baseline reaches 70.11~mIoU. 
When enabling all-stage cross-attention without LR supervision ($\alpha{=}0$), performance increases to 70.30~mIoU, and further rises to 71.40~mIoU once auxiliary LR supervision is added ($\alpha{=}0.5$). 
This corresponds to a gain of +1.29~mIoU over the re-trained Swin-B baseline, indicating that the improvement comes from explicit context-aware fusion.
We also verified that our Swin-B reproduction closely matches the official FLAIR-HUB RGB results on the test set (see \S~\ref{sec:quantitative_results}), which validates our implementation and training setup.

\paragraph{Stage-wise fusion.}
We compare cross-attention applied only at the first stage, only at the last stage, and at all four encoder stages. 
Using cross-attention exclusively at the last stage (Stage-4 only) already provides a strong improvement over the single-stream baseline (71.32 vs.\ 70.11~mIoU), confirming that injecting LR context at a high semantic level is beneficial. 
Relying on the first stage alone is less effective (69.89~mIoU), suggesting that early low-level interaction is not sufficient by itself. 
Our full CASWiT-Base model, which performs stage-wise fusion at all levels, achieves the best overall result (71.40~mIoU), indicating that combining early and late cross-scale interactions yields the most balanced trade-off between fine detail and global coherence.

\paragraph{Auxiliary LR supervision.}
We vary the auxiliary loss weight $\alpha$ from 0 to 0.5 in Table~\ref{tab:ablation}, and additionally observe that performance remains robust for moderate values around $\alpha=0.5$. In particular, values in the range $\alpha \in [0.3, 0.7]$ yield comparable validation performance (70.7--71.3 mIoU), whereas more extreme settings reduce performance (below 68.6 mIoU for $\alpha=0.1$ and $\alpha=0.9$). Comparing the all-stage variants, adding LR supervision ($\alpha{=}0.5$) improves mIoU from 70.30 to 71.40 and also leads to smoother validation curves (not shown), suggesting that lightweight LR guidance regularizes the shared representation and facilitates optimization.

\paragraph{Gating mechanism.}
We evaluate the optional learned gate $g_s$ used to scale the cross-attention residuals. 
With all-stage fusion and $\alpha{=}0$, enabling gating yields 70.15~mIoU, slightly below the ungated counterpart (70.30~mIoU). 
We did not observe consistent benefits in terms of stability or final accuracy, and therefore keep gating disabled in the main configuration (equivalent to $\gamma_s = 1$).

\paragraph{Model size.}
We additionally evaluate a lighter variant, CASWiT-Tiny, which uses the same cross-attention design but a reduced-capacity backbone. 
CASWiT-Tiny reaches 70.91~mIoU, only 0.49~mIoU below CASWiT-Base despite its smaller parameter budget. 
This indicates that the proposed fusion strategy can still provide benefits in a more compact setting.

\paragraph{Self-supervised pretraining.}
Finally, we compare models trained from scratch to those initialized with our SimMIM-style pretraining on SWISSIMAGE. 
On the FLAIR-HUB validation set, pretraining increases performance from 71.40 to 71.55~mIoU (+0.15), and we observe larger gains on the test set (see \S~\ref{sec:experiments}). 
Qualitatively, the pretrained model produces sharper boundaries and more coherent large structures, indicating that masked reconstruction on large-scale orthoimagery exposes the network to structural cues beneficial for UHR semantic segmentation. 
The centered LR masking is designed so that the model cannot rely on trivial cross-scale copying from the LR region spatially aligned with the HR crop, and must instead exploit the surrounding LR context, which better matches the downstream role of cross-attention.

\begin{table}[t]
\centering
\scriptsize
\setlength{\tabcolsep}{6pt}
\begin{tabular}{l | c c | c c}
\toprule
Variant & Cross-Attn & $\alpha$ & mIoU (\%)↑ & mF1 (\%)↑ \\
\midrule
Baseline (no CA) & None & 0.0 & 70.11 & 81.72 \\
All-stage + gating on &  (All, gated) & 0.0 & 70.15 & 81.78  \\
All-stage &  (All) & 0.0 & 70.30 & 81.89 \\
Stage-1 fusion only &  (Stage 1) & 0.5 &  69.89 & 81.56  \\
Last-stage fusion only &  (Stage 4) & 0.5 & 71.32 & 82.56 \\
CASWiT-Tiny  &  (All) & 0.5 & 70.91 & 82.24 \\
CASWiT-Base &  (All) & 0.5 & 71.40 & 82.62 \\
CASWiT-Base-SSL &  (All) & 0.5 & 71.55  & 82.78 \\
\bottomrule
\end{tabular}
\caption{Ablation study on the FLAIR-HUB validation set. Each component is varied independently; CA denotes cross-attention and $\alpha$ the LR supervision weight. All models use a Swin-B backbone, except CASWiT-Tiny, and are trained without augmentation.}
\label{tab:ablation}
\end{table}

\begin{figure*}
    \centering
    \includegraphics[width=1\linewidth]{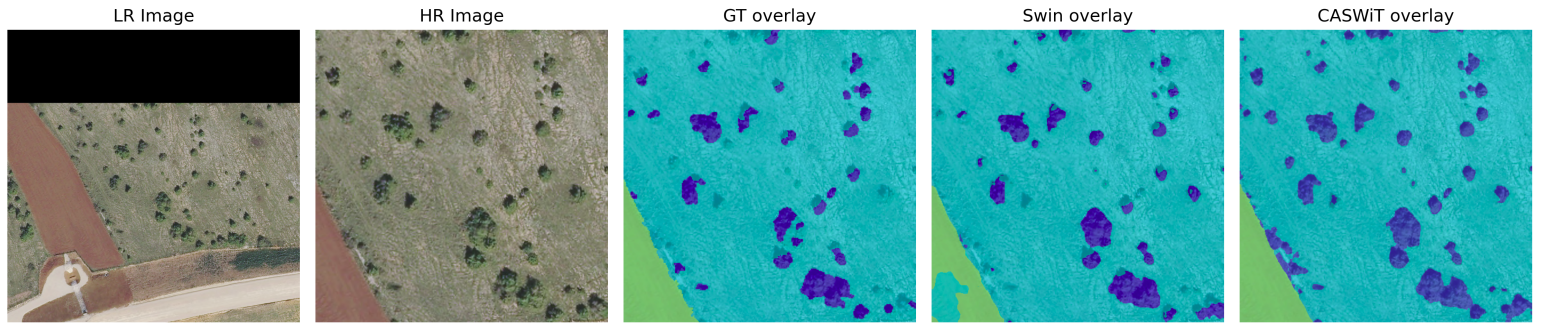}
    \caption{Qualitative comparison on FLAIR-HUB. From left to right: LR image (note the missing band at the top), HR crop, ground-truth overlay, RGB baseline (Swin-B + UPerNet) overlay, and CASWiT overlay. CASWiT better recovers small vegetation patches and yields sharper boundaries, while reducing false positives on bare soil and road areas. Despite the LR artifact, CASWiT remains stable.}
    \label{fig:qual_comp_flairhub}
\end{figure*}

\begin{figure*}
    \centering
    \includegraphics[width=1\linewidth]{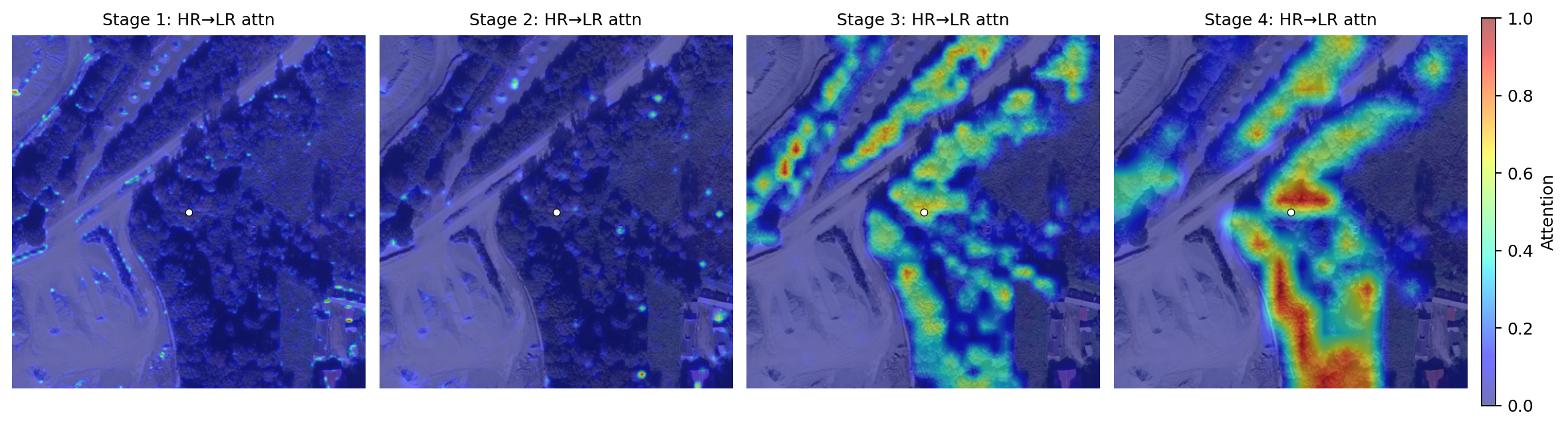}
    \caption{Cross-attention maps after self-supervised pretraining and supervised fine-tuning. Visualization of HR-to-LR cross-attention at each encoder stage of CASWiT. The queried HR pixel is marked by a white dot, and attention weights are reprojected onto the LR token grid and overlaid on the LR image.}
    \label{fig:cross_attn_ssl_finetune}
\end{figure*}

\subsection{Qualitative Analysis}\label{sec:qualitative}
We provide qualitative visualizations to further illustrate the behavior of CASWiT and the role of cross-scale context injection.

\paragraph{Segmentation results on FLAIR-HUB.}
Fig.~\ref{fig:qual_comp_flairhub} shows a representative example from the FLAIR-HUB test set comparing CASWiT-Base-SSL + UPerNet to the RGB Swin-B + UPerNet baseline (more examples are provided in the supplementary). Our model produces cleaner boundaries and better preserves fine structures, such as narrow roads and building outlines. It also reduces semantic bleeding between adjacent classes.

\paragraph{Cross-attention visualization.}
To understand how context propagates across scales, we visualize the attention maps from HR queries to LR keys at different encoder stages (Fig.~\ref{fig:cross_attn_ssl_finetune} and additional examples in the supplementary). Early stages (stages 1-2) tend to concentrate on fine local structures and boundaries, whereas later stages (stages 3-4) attend more broadly to the semantic layout, such as roads, vegetation, and water. This progressive behavior suggests that multi-stage cross-attention integrates contextual information across different levels of the hierarchy.

\paragraph{Self-supervised reconstruction.}
Fig.~\ref{fig:mae_figure} illustrates the masked reconstruction process during self-supervised pretraining. Random HR masks (75\%) and centered LR masks (50\%) are applied jointly; the network must reconstruct missing HR pixels using both visible HR content and surrounding LR context. CASWiT successfully recovers fine-grained textures and object geometry, indicating that the dual-stream fusion effectively learns cross-scale correspondences.

\section{Conclusion}\label{sec:conclusion}

%\subsection{Conclusion and Limitations}
%We introduced \textbf{CASWiT}, a cross-attentive dual-branch backbone for ultra-high-resolution RGB segmentation that fuses HR detail with LR context via lightweight, stage-wise cross-attention. We also proposed an RGB-only FLAIR-HUB-UHR evaluation protocol that leverages geospatial structure to preserve long-range context while remaining comparable to the original patch-based setting of FLAIR-HUB~\cite{flairhub}. On this benchmark, CASWiT establishes a strong RGB-only reference, reaching 66.37 mIoU with a SegFormer decoder and improving boundary quality by +4.33 mBIoU over the retrained Swin-B + UPerNet baseline. Notably, this best RGB-only variant also exceeds the 65.9 mIoU reported on FLAIR-HUB by the multimodal Maestro configuration~\cite{maestro}. Beyond remote sensing, CASWiT also transfers effectively to medical UHR segmentation benchmarks, reaching 86.5 mIoU on ISIC~\cite{isic} and 90.7 mIoU on CRAG~\cite{CRAG_dataset}. Taken together, these results indicate that the gains stem primarily from explicit cross-scale fusion and are further reinforced by SimMIM-style pretraining on large-scale orthophotos.

\subsection{Conclusion and Limitations}
We introduced \textbf{CASWiT}, a cross-attentive dual-branch backbone for ultra-high-resolution RGB segmentation that fuses HR detail with LR context via lightweight, stage-wise cross-attention. We also proposed an RGB-only FLAIR-HUB-UHR evaluation protocol that leverages geospatial structure while remaining comparable to the original patch-based setting of FLAIR-HUB~\cite{flairhub}. On this benchmark, CASWiT reaches 66.37 mIoU with a SegFormer decoder and improves boundary quality by +4.33 mBIoU over the retrained Swin-B + UPerNet baseline. This best RGB-only variant also exceeds the 65.9 mIoU reported on FLAIR-HUB by the multimodal Maestro configuration~\cite{maestro}. On URUR, CASWiT reaches 49.2 mIoU, improving over prior UHR-specific methods under the official protocol. CASWiT also transfers effectively to medical UHR segmentation benchmarks, reaching 86.5 mIoU on ISIC~\cite{isic} and 90.7 mIoU on CRAG~\cite{CRAG_dataset}. Overall, these results suggest that the gains stem primarily from explicit cross-scale fusion and are further reinforced by SimMIM-style pretraining on large-scale orthophotos. A current limitation is that CASWiT still relies on a dual-branch design, which increases architectural and computational complexity relative to simpler single-stream baselines.

\subsection{Perspectives}
Our core contribution is the backbone: CASWiT delivers stronger, context-aware features while remaining compatible with different decoder heads. A natural next step is to extend the evaluation of CASWiT to broader high-resolution vision settings. In particular, competitive results obtained with a CASWiT-based model in the NTIRE 2026 Remote Sensing Infrared Image Super-Resolution challenge suggest that improved context-aware feature extraction is also a promising direction for super-resolution~\cite{ntire26rsirsr}.

\section{Acknowledgements}
We would like to thank Fabien Délèze for his careful proofreading, and the ESO team for their support. We also thank Shanci Li for his valuable assistance with the dataset, as well as Amir Zamir and his team for their constructive feedback and insightful discussions as part of the Visual Intelligence course. This research was supported by the Canton of Vaud and the INSIT Institute at HEIG-VD.

{
    \small
    \bibliographystyle{ieeenat_fullname}
    \bibliography{main}
}
 
\clearpage
\setcounter{page}{1}
\maketitlesupplementary

\section{URUR: illustrative annotation mismatch}

\begin{strip}
\raggedright
\justifying

\begin{center}
\captionsetup{type=figure}

\begin{tabular}{ccc}
\includegraphics[width=0.3\linewidth]{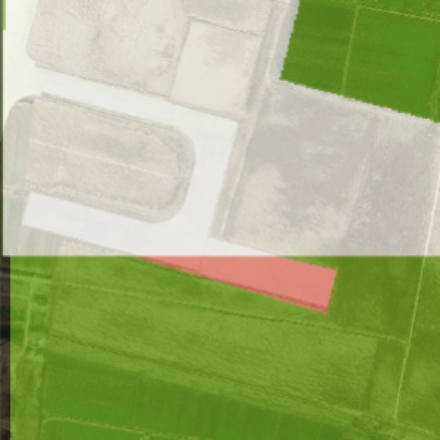} &
\includegraphics[width=0.3\linewidth]{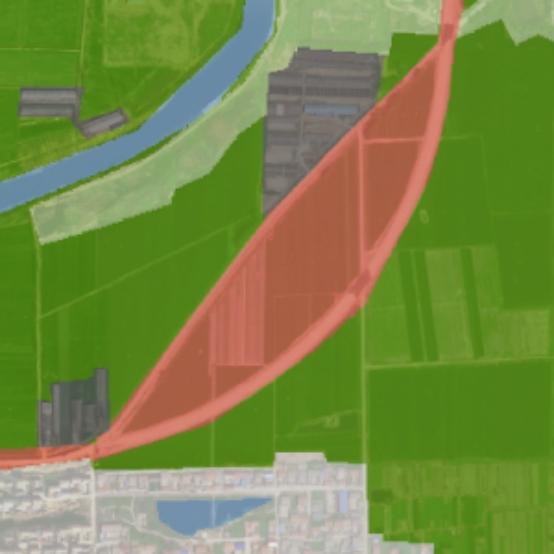} &
\includegraphics[width=0.3\linewidth]{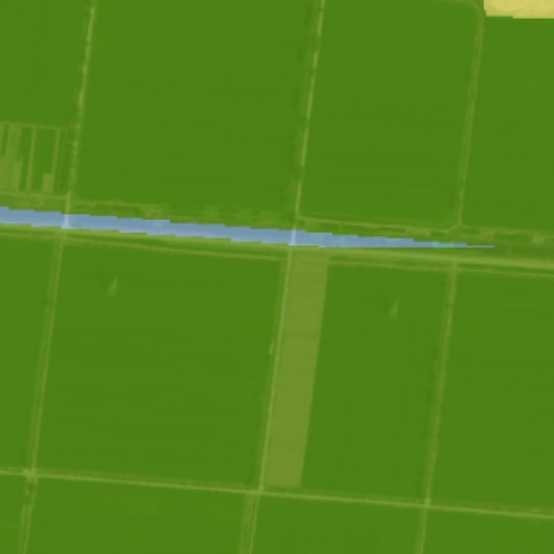} \\[0.3cm]

\includegraphics[width=0.3\linewidth]{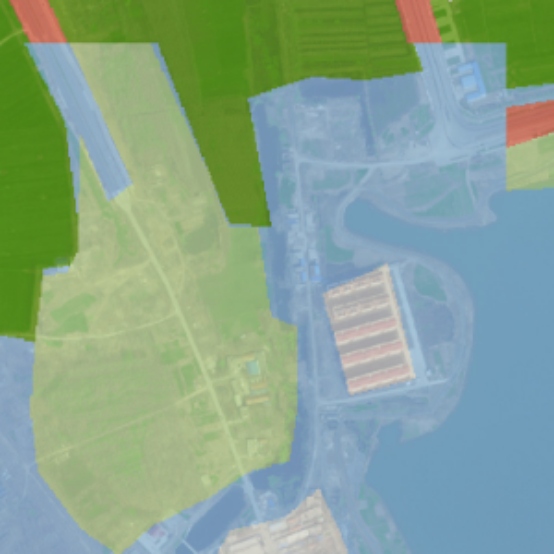} &
\includegraphics[width=0.3\linewidth]{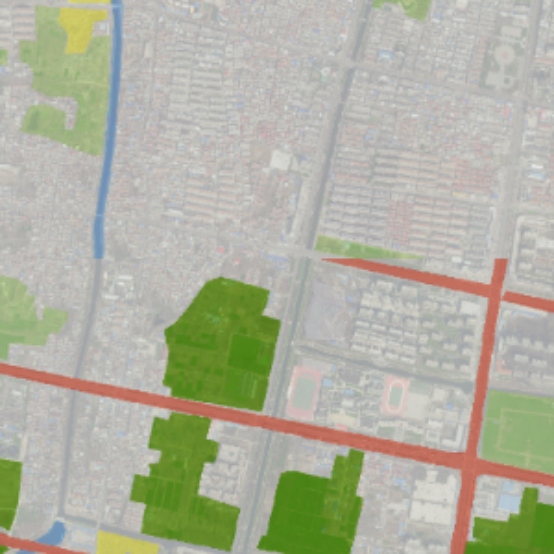} &
\includegraphics[width=0.3\linewidth]{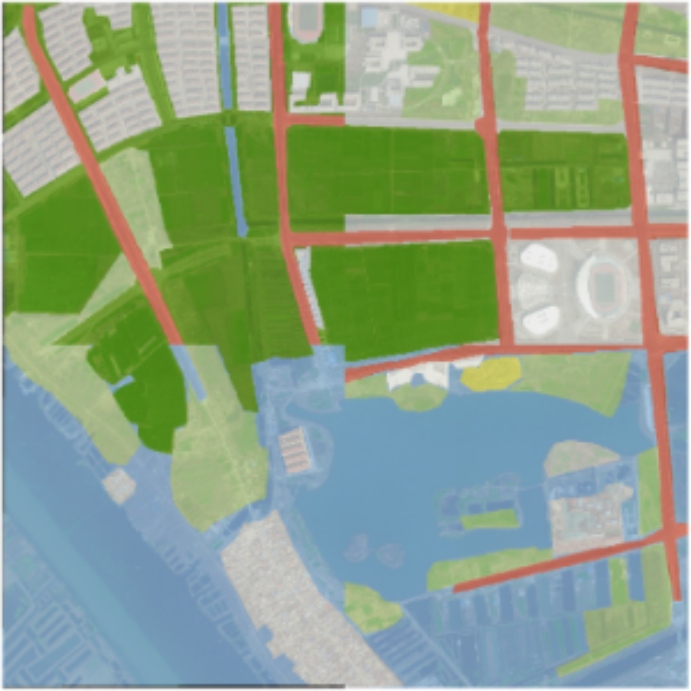} \\
\end{tabular}

\caption{Example where the provided mask (overlaid) locally diverges from the RGB content; such cases are occasional but can affect evaluation metrics. Visible classes include:
\textcolor[rgb]{0,0,0}{\textbf{others}},
\textcolor[rgb]{0.90,0.90,0.90}{\textbf{building}},
\textcolor[rgb]{0.39,0.39,0.39}{\textbf{greenhouse}},
\textcolor[rgb]{0.78,0.90,0.63}{\textbf{woodland}},
\textcolor[rgb]{0.37,0.64,0.03}{\textbf{farmland}},
\textcolor[rgb]{1.0,1.0,0.39}{\textbf{bareland}},
\textcolor[rgb]{0.59,0.78,0.98}{\textbf{water}},
\textcolor[rgb]{0.94,0.39,0.31}{\textbf{road}}.
}
\label{fig:urur_mismatch_supp}
\end{center}

\newpage

\section{Qualitative analysis on FLAIR-HUB}
\label{app:visual_comparison}
\begin{center}
\captionsetup{type=figure}
    \includegraphics[width=1\linewidth]{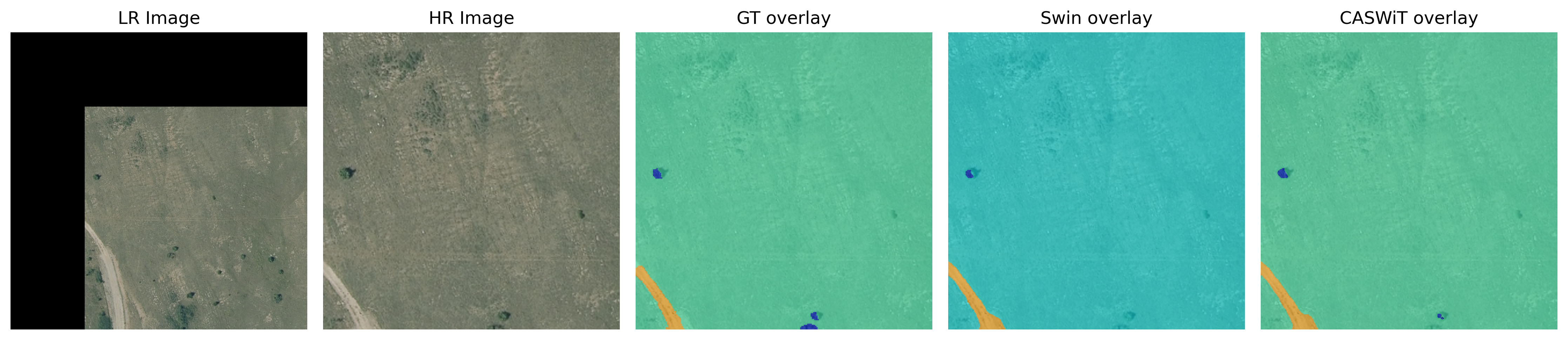}
    \includegraphics[width=1\linewidth]{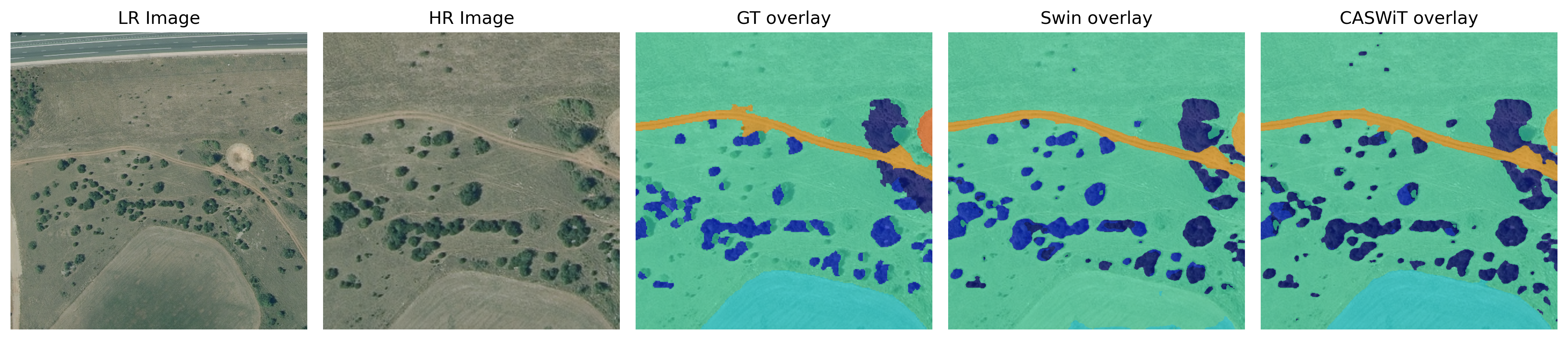}
    \includegraphics[width=1\linewidth]{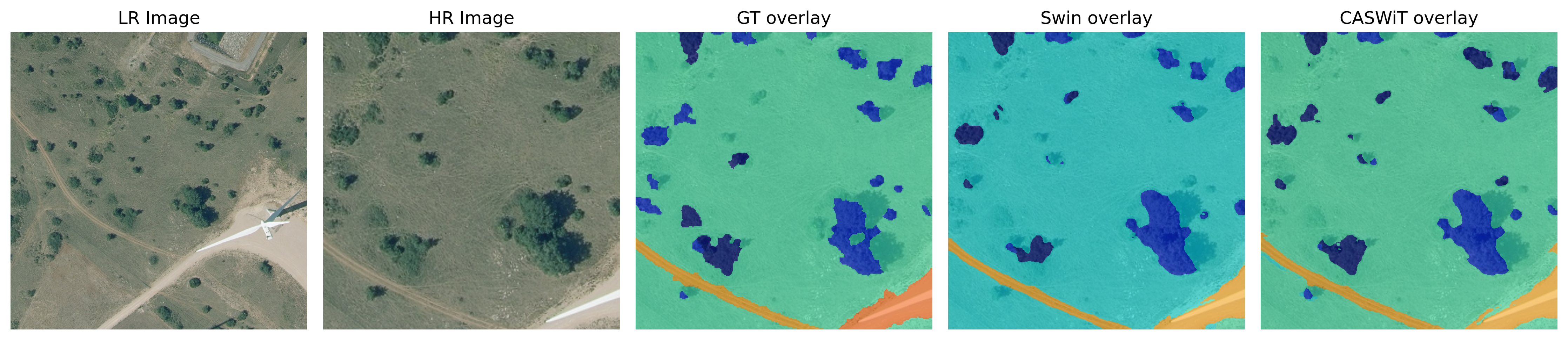}
    \includegraphics[width=1\linewidth]{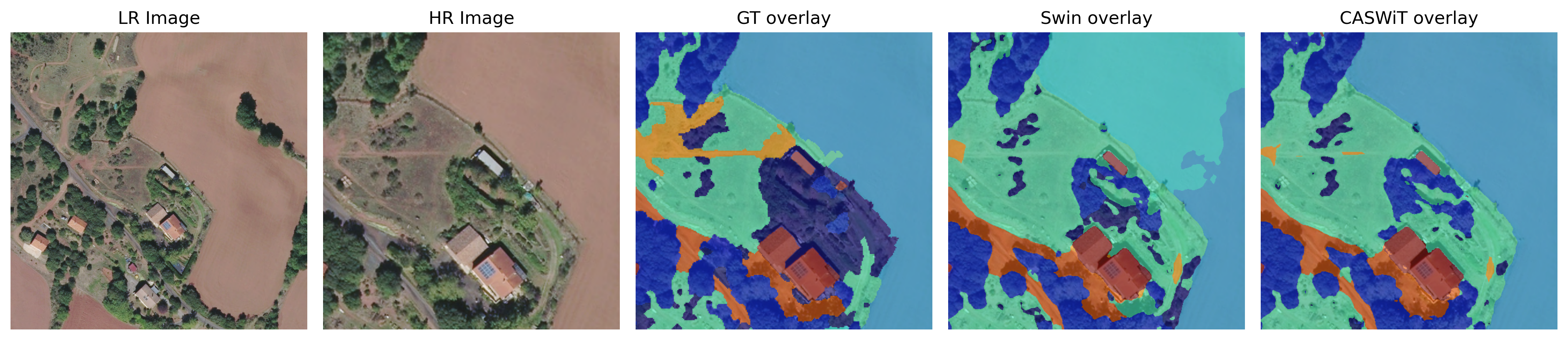}
    \includegraphics[width=1\linewidth]{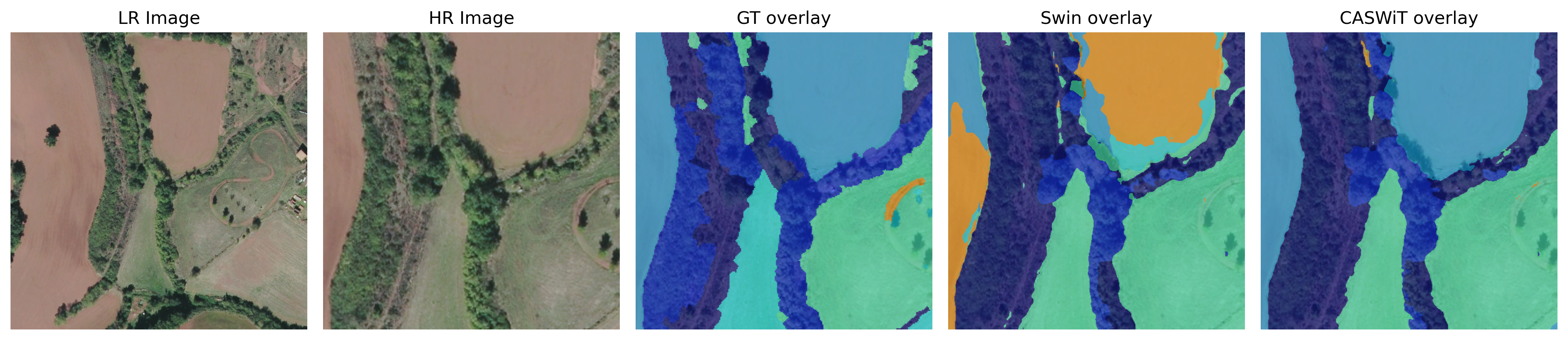}
    \includegraphics[width=1\linewidth]{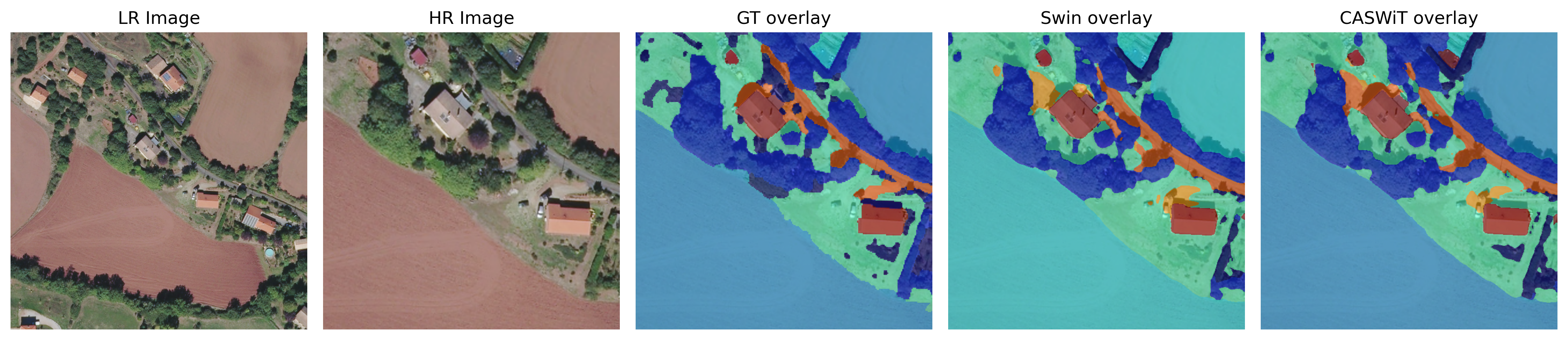}
    \includegraphics[width=1\linewidth]{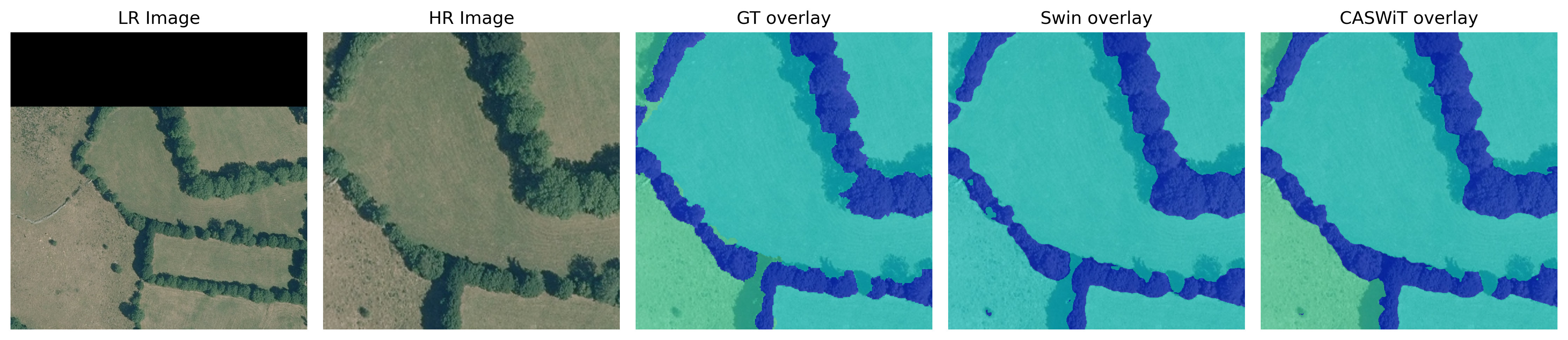}
    \includegraphics[width=1\linewidth]{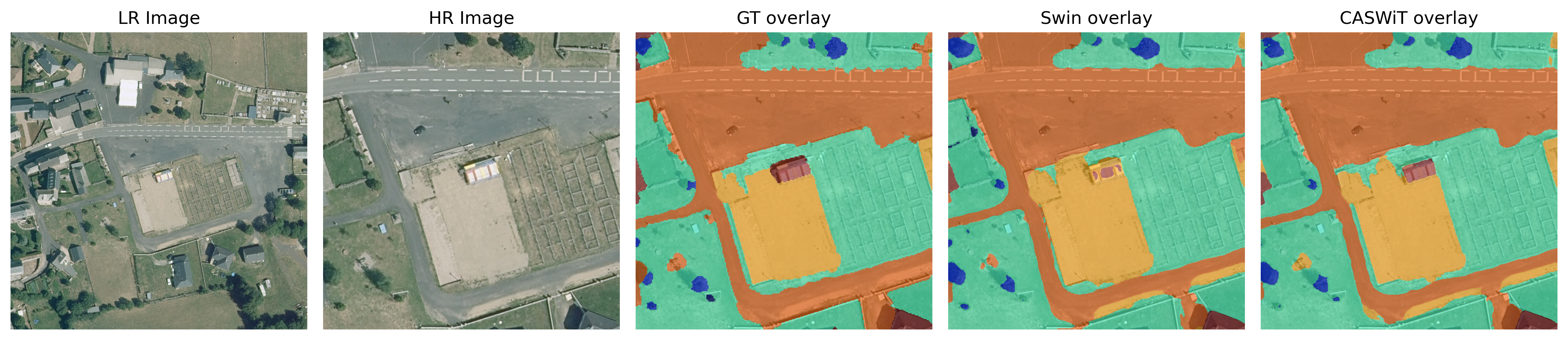}
 \caption{Comparison of LR/HR images, ground truth overlays, Swin Base predictions, and CASWiT predictions on eight test patches.}
 \label{fig:comparison_overlays}
\end{center}

\section{Supplementary results (IoUs)}

\begin{center}
\captionsetup{type=table}
\begin{tabular}{l|c|cc}
\toprule
Class & Swin-B~\cite{flairhub} & CASWiT-B-SSL-aug + UPer & CASWiT-B-SSL-aug + SegF\\
\midrule
Building              & 83.77 & 85.47 & 85.27 \\
Greenhouse            & 77.89 & 79.46 & 80.67 \\
Swimming pool         & 61.59 & 62.12 & 60.48 \\
Impervious surface    & 75.03 & 76.78 & 76.94 \\
Pervious surface      & 56.97 & 58.86 & 58.73 \\
Bare soil             & 65.21 & 66.95 & 67.79 \\
Water                 & 90.08 & 90.65 & 90.35 \\
Snow                  & 67.77 & 66.59 & 75.95 \\
Herbaceous vegetation & 52.85 & 55.07 & 54.39 \\
Agricultural land     & 56.53 & 60.38 & 60.63 \\
Plowed land           & 37.34 & 38.20 & 38.49 \\
Vineyard              & 78.88 & 80.71 & 80.71 \\
Deciduous             & 70.07 & 71.47 & 70.89 \\
Coniferous            & 58.95 & 62.89 & 62.73 \\
Brushwood             & 30.97 & 31.79 & 31.61 \\
\midrule
mIoU                  & 64.05 & 65.83 & \textbf{66.37} \\
\bottomrule
\end{tabular}
\caption{Per-class IoU (\%) on the FLAIR-HUB-RGB test set for Swin-B and our CASWiT variants.}
\label{tab:flairhub_per_class}
\end{center}

\begin{center}
\captionsetup{type=table}
\begin{tabular}{l|cc|cc}
\toprule
Class & WSDNET~\cite{WSDNET} & Boosting Dual-Stream~\cite{boosting}& CASWiT-B-SSL-aug + UPer & CASWiT-B-SSL-aug + SegF \\
\midrule
Others    & - & - & 0.00  & 0.00 \\
Building  & - & - & 75.07 & 74.42 \\
Farmland  & - & - & 79.19 & 79.31 \\
Greenhouse& - & - & 46.51 & 46.50 \\
Woodland  & - & - & 52.10 & 51.79 \\
Bareland  & - & - & 31.64 & 32.31 \\
Water     & - & - & 54.90 & 55.86 \\
Road      & - & - & 53.33 & 53.67 \\
\midrule
mIoU      & 46.9 & 48.2 & 49.1 & \textbf{49.2} \\
\bottomrule
\end{tabular}
\caption{Per-class IoU (\%) on the URUR dataset test set for our CASWiT-Base variants.}
\label{tab:urur_per_class}
\end{center}

%\newpage
\section{Dataset FLAIR-HUB merge}\label{app:dataset_blakc_filling}

\begin{center}
\captionsetup{type=figure}
    \centering
    \includegraphics[width=0.9\linewidth]{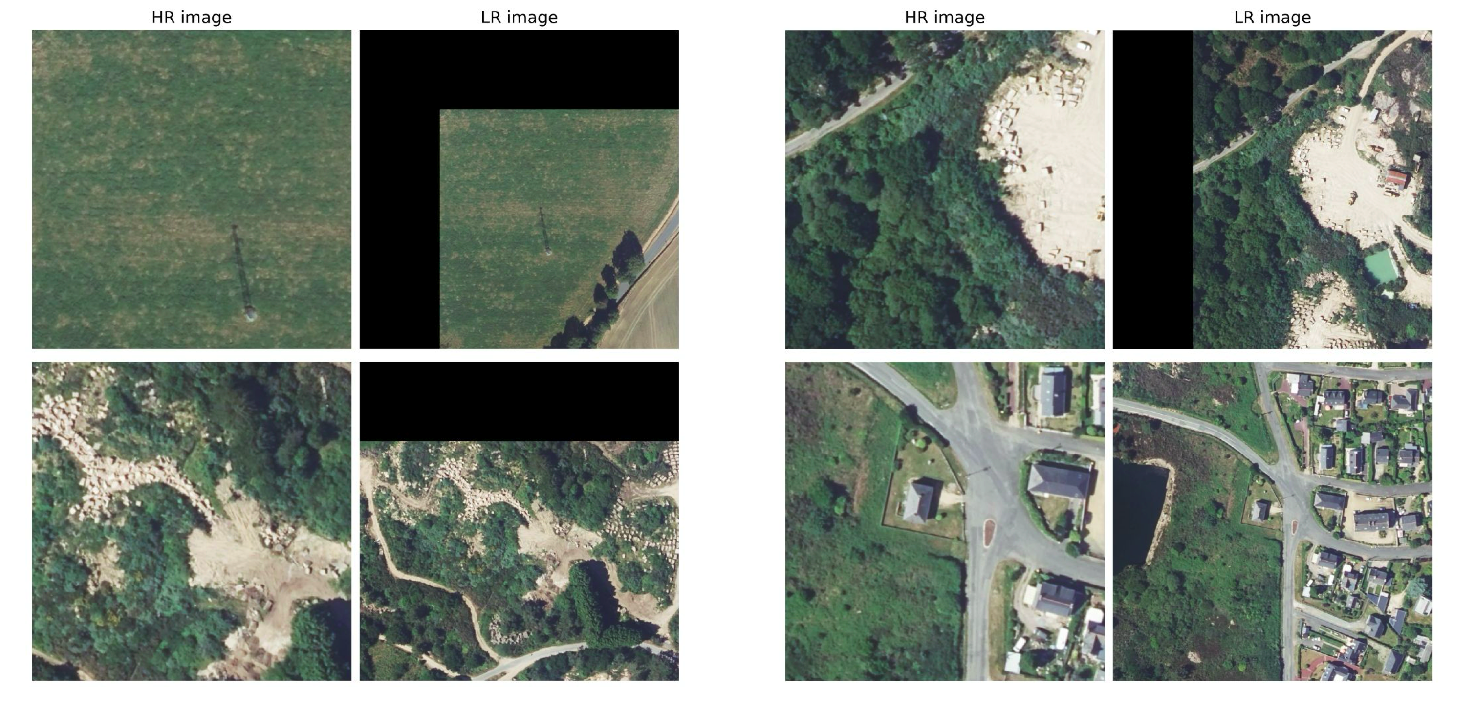}
  \caption{Examples of data pre-processing, on the left are the HR patches and on the right are the merged patches obtained from the available neighbors.}
  \label{fig:borders_LR}
\end{center}

\newpage

\section{SSL results}\label{app:mae}
\begin{center}
\captionsetup{type=figure}
  \centering
  \includegraphics[width=1\textwidth]{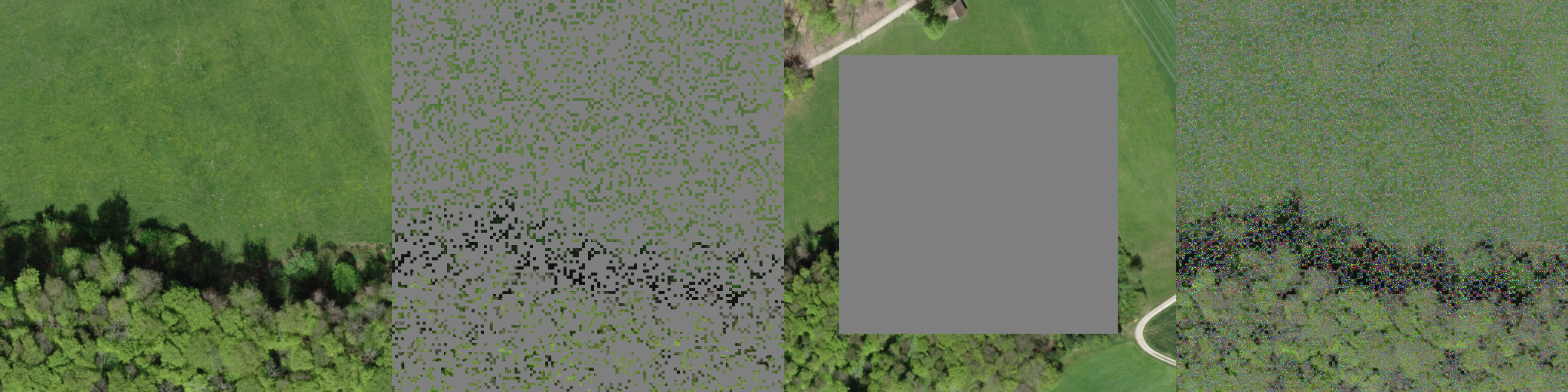}
  \includegraphics[width=1\textwidth]{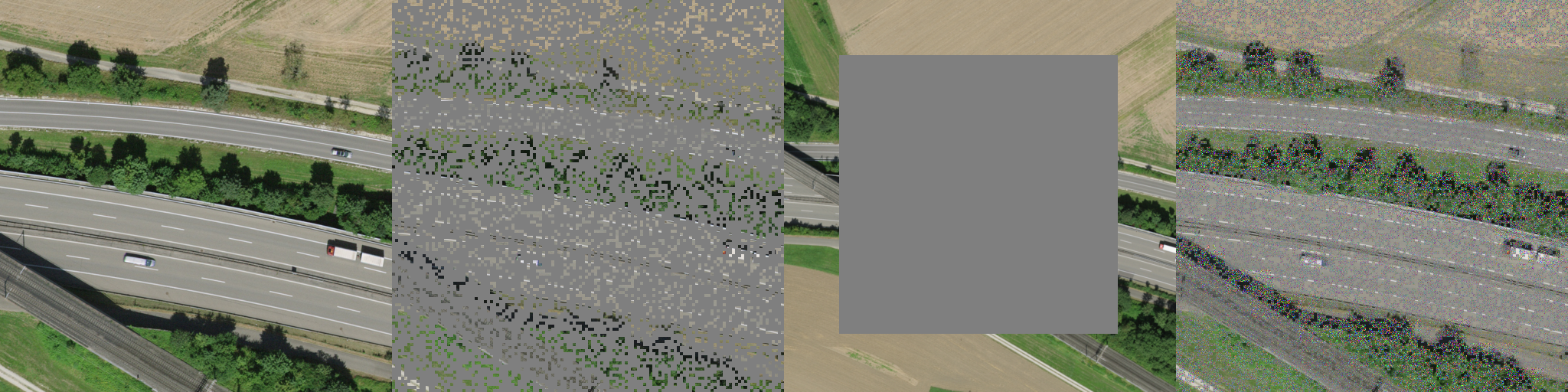}
  \includegraphics[width=1\textwidth]{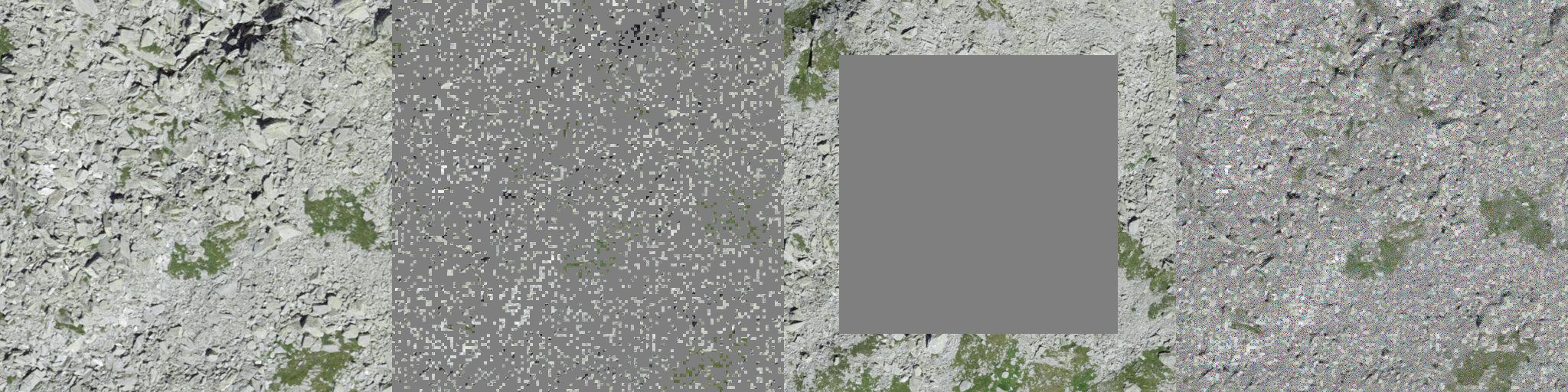}
  \includegraphics[width=1\textwidth]{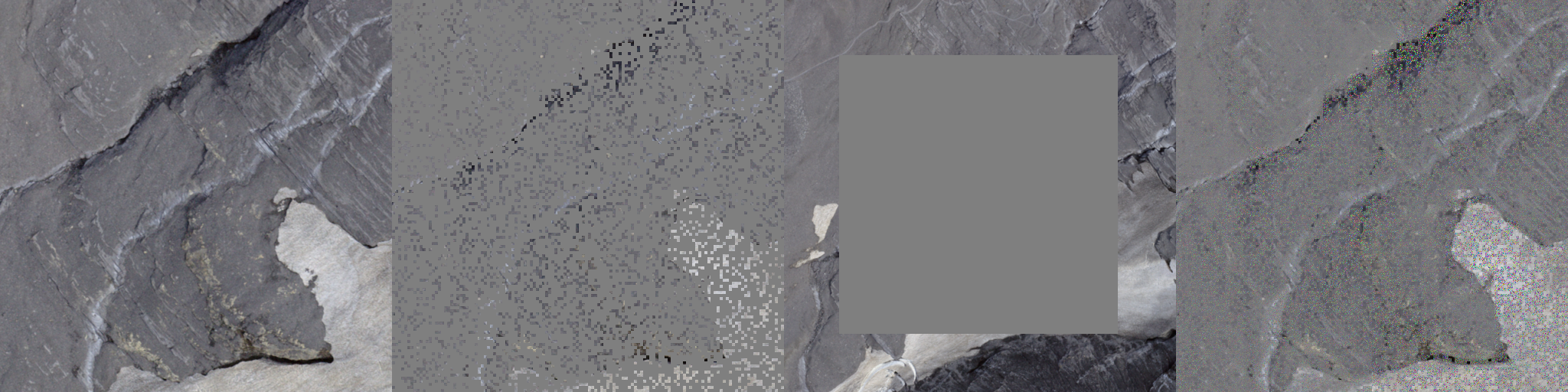}
  \includegraphics[width=1\textwidth]{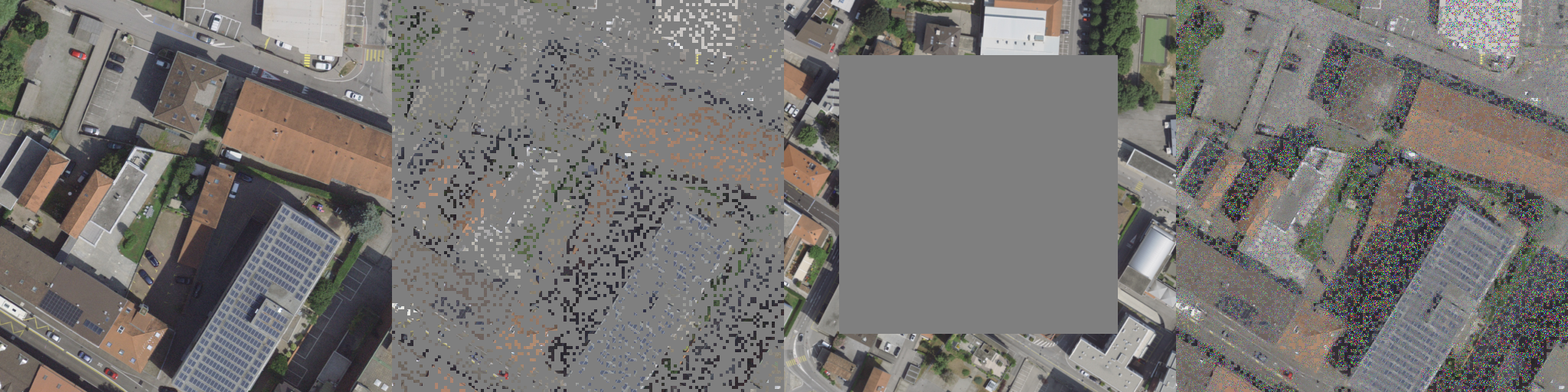}
  \includegraphics[width=1\textwidth]{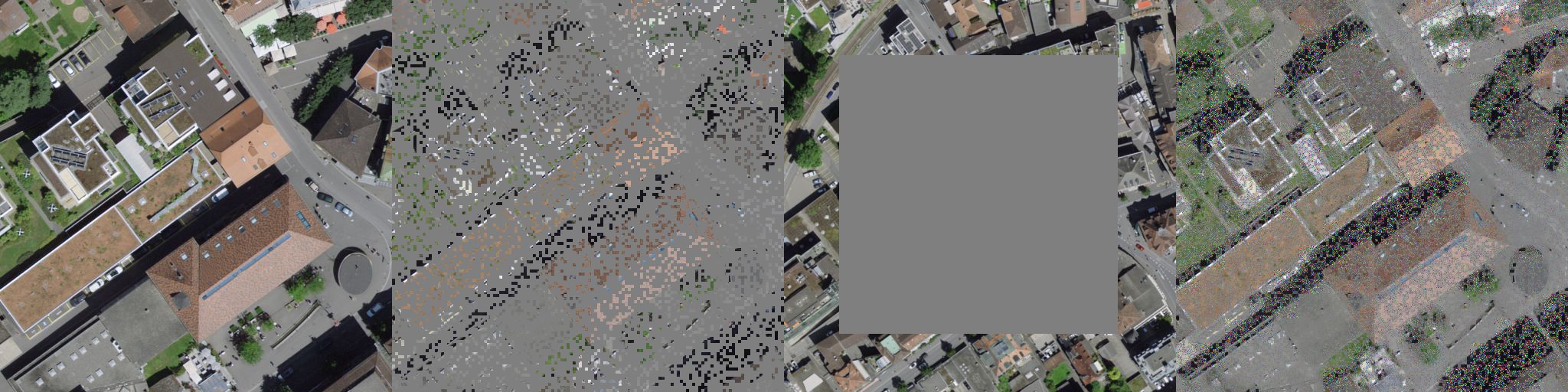}
  \includegraphics[width=1\textwidth]{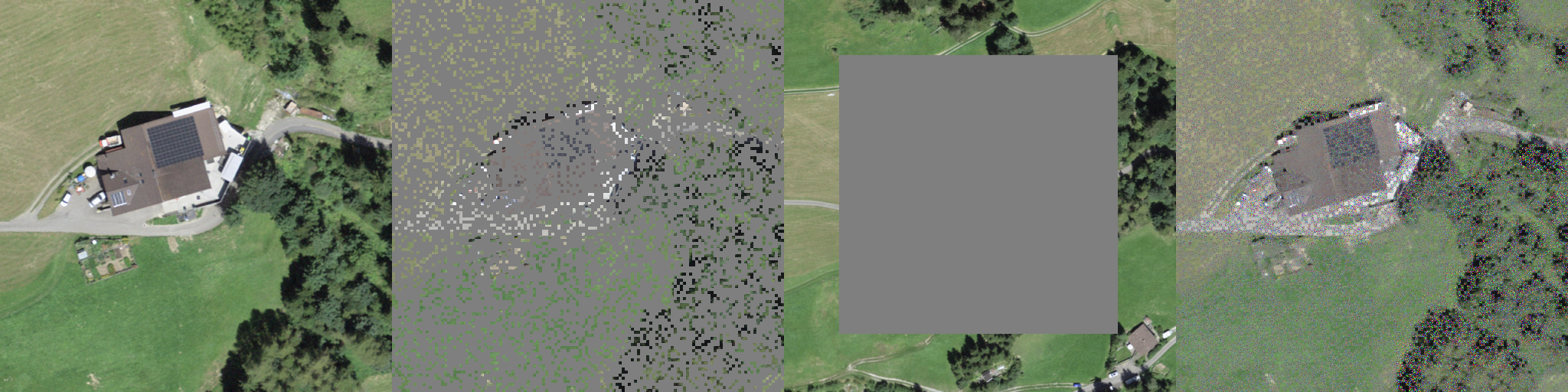}
  \caption{Self-supervised SimMIM-style inference results on the CASWiT-Base architecture. Each row (left to right) shows: original high-resolution image, high-resolution image with random masking, low-resolution image with central masking, and the reconstruction of the high-resolution image.}
  \label{fig:mae_ccast_appendix}
\end{center}

\newpage
\section{Cross-attention visualization}

\begin{center}
\captionsetup{type=figure}
  \centering
  \includegraphics[width=1\textwidth]{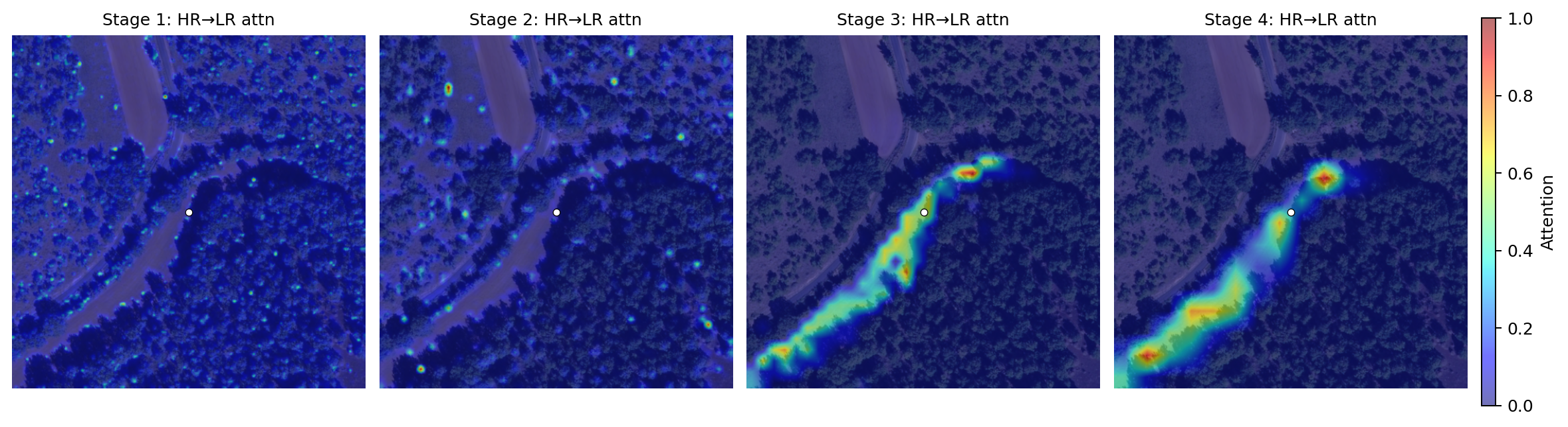}
  \includegraphics[width=1\textwidth]{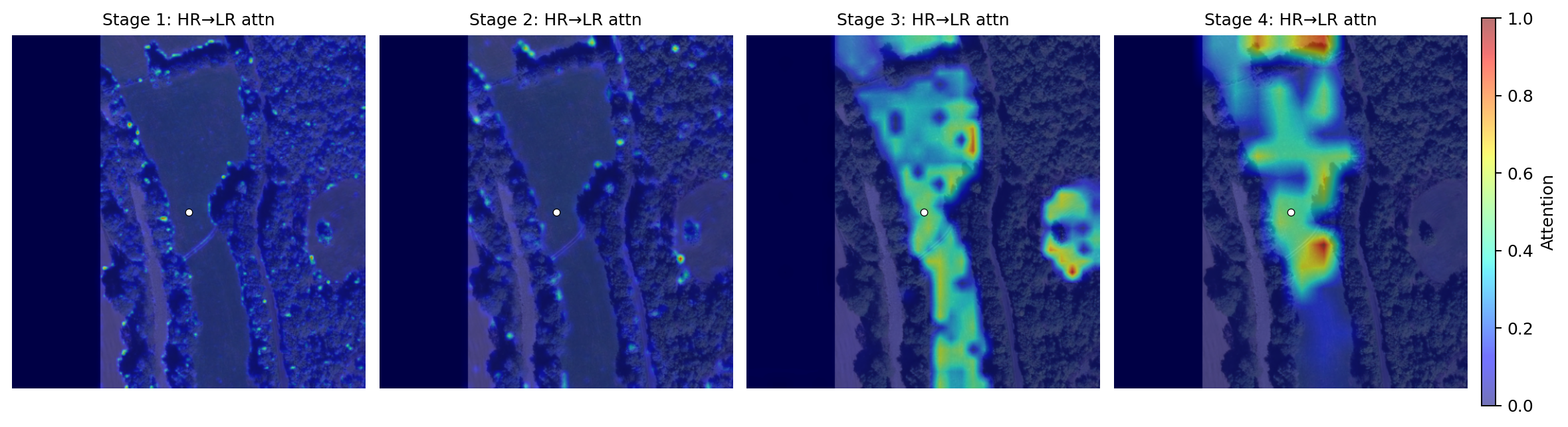}
  \includegraphics[width=1\textwidth]{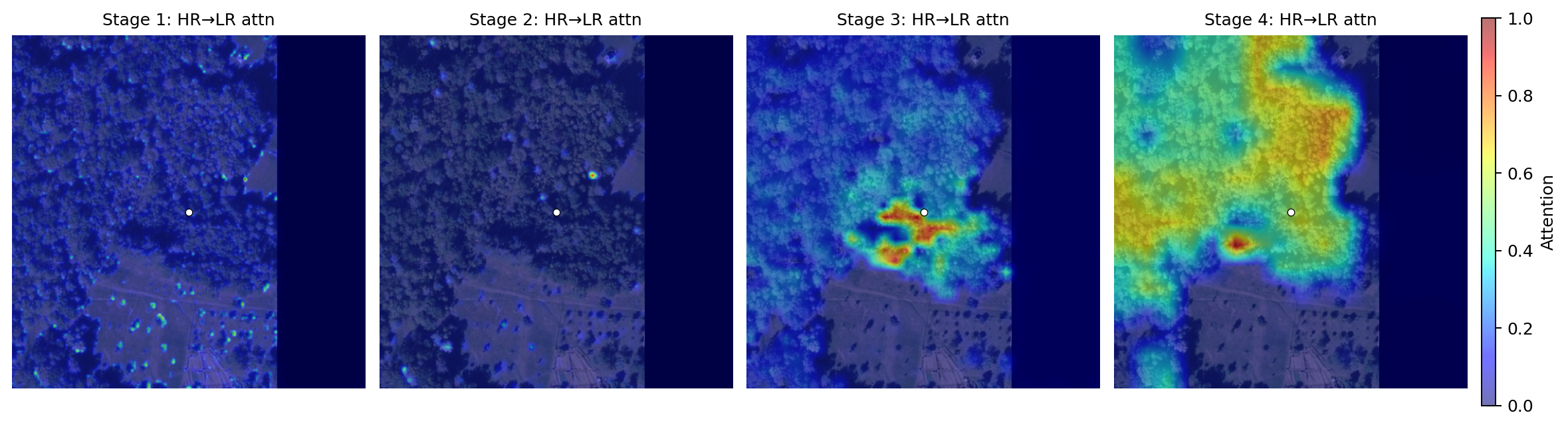}
  \includegraphics[width=1\textwidth]{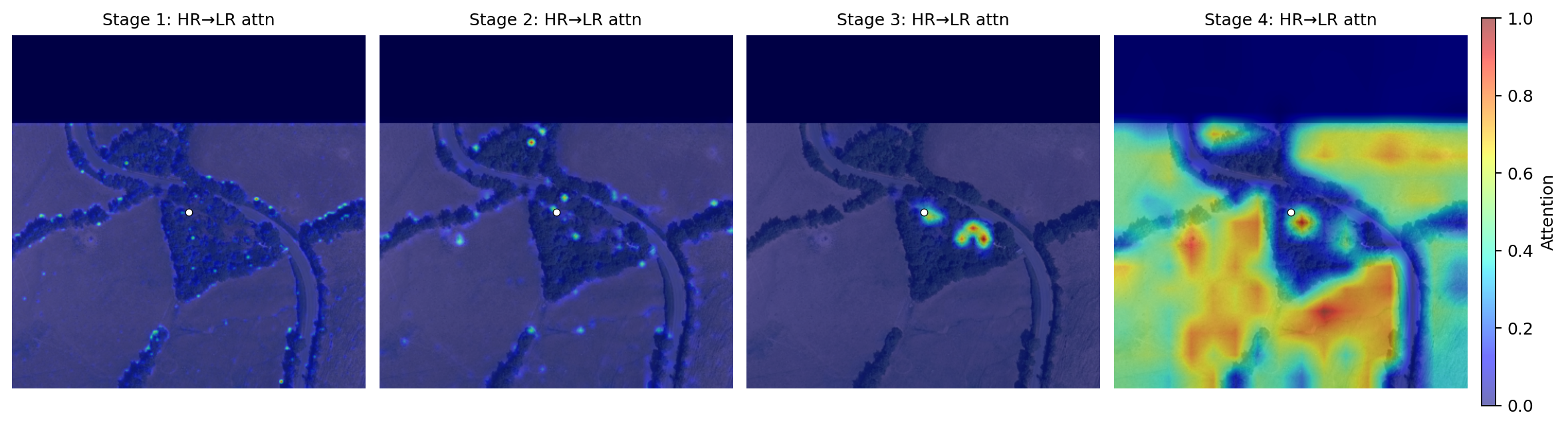}
  \caption{Visualization of cross-attention maps for each stage of the model on four test patches.}
  \includegraphics[width=1\textwidth]{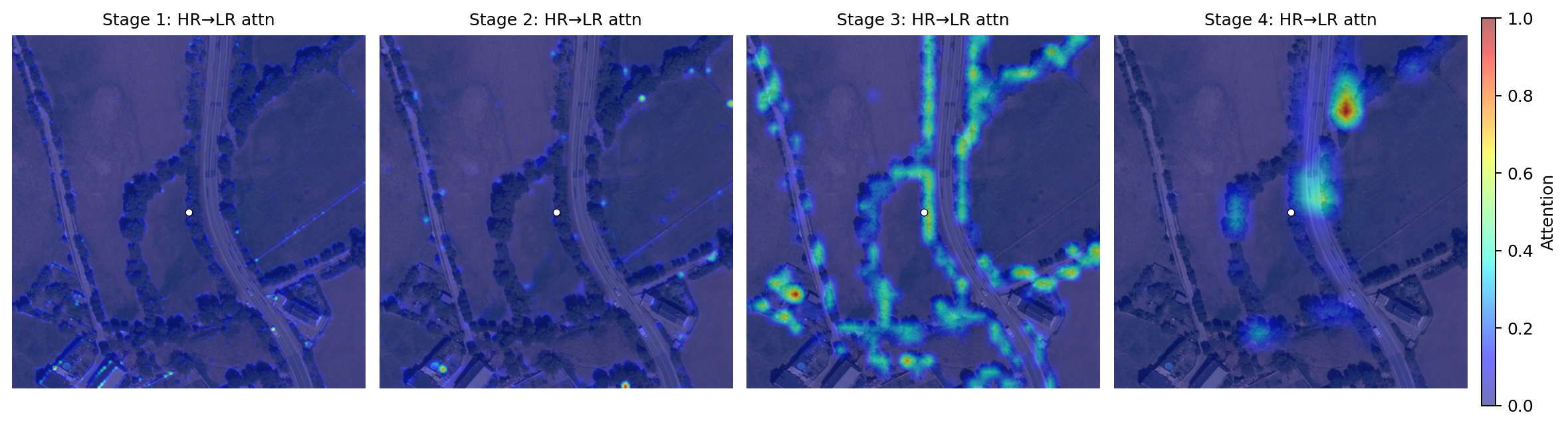}
  \label{fig:xattn_appendix}
\end{center}
\end{strip}

\end{document}